\documentclass[]{elsarticle}

\usepackage{amssymb}

\usepackage{url}
\usepackage{booktabs}
\usepackage{lscape}
\usepackage{hyperref}
\usepackage{caption}
\usepackage{subcaption}
\usepackage{todonotes}
\usepackage{makecell}
\usepackage{xspace}
\usepackage{multirow}
\usepackage{bm}
\usepackage{tikz,tikzpeople}
\usetikzlibrary{arrows,shapes}
\usepackage{amsmath}
\usepackage{algorithm}
\usepackage[noend]{algorithmic}

\DeclareMathOperator*{\argmax}{argmax}
\newcommand{\ltlf}{\texttt{LTL$_f$}\xspace}
\newcommand{\ltl}{\texttt{LTL}\xspace}
\newcommand{\lnext}{\ensuremath{\mathbf{X}}}
\newcommand{\lfuture}{\ensuremath{\mathbf{F}}}
\newcommand{\luntil}{\ensuremath{\mathbf{U}}}
\newcommand{\lglobally}{\ensuremath{\mathbf{G}}}
\newcommand\act[1]{\texttt{#1}}
\newcommand\attr[1]{\texttt{\textit{#1}}}
\newcommand\constr[1]{\textsc{#1}}
\newcommand{\tup}[1]{\langle #1\rangle}  
\newcommand{\declare}{\textsc{declare}\xspace}
\newcommand{\Log}{\ensuremath{\mathcal{L}}\xspace}

\newcommand\revision[1]{{\color{black}{#1}}}
\newcommand\secRevision[1]{{\color{black}{#1}}}
\newtheorem{definition}{Definition}

\begin{document}

\begin{frontmatter}

\title{Outcome-Oriented Prescriptive Process Monitoring Based on Temporal Logic Patterns}

\author[unibz]{Ivan Donadello\corref{cor1}}
\ead{ivan.donadello@unibz.it}
\cortext[cor1]{Corresponding author}
\affiliation[unibz]{organization={Faculty of Computer Science, Free University of Bozen-Bolzano},
            addressline={Piazza Domenicani 3}, 
            city={Bolzano},
            postcode={39100},
            country={Italy}}

\author[unitn,fbk]{Chiara Di Francescomarino}
\ead{ivan.donadello@unibz.it}
\affiliation[unitn]{organization={Department of Information Engineering and Computer Science, University of Trento},
            addressline={Via Sommarive 9}, 
            city={Trento},
            postcode={38123}, 
            country={Italy}}
\affiliation[fbk]{organization={Center for Digital Health \& Wellbeing, Fondazione Bruno Kessler},
            addressline={Via Sommarive 18}, 
            city={Trento},
            postcode={38123}, 
            country={Italy}}

\author[unibz]{Fabrizio Maria Maggi}
\ead{maggi@inf.unibz.it}

\author[unibz]{Francesco Ricci}
\ead{Francesco.Ricci@unibz.it}

\author[Visioncraft]{Aladdin Shikhizada}
\ead{aladdin.shikhizada@visioncraft.ee}
\affiliation[Visioncraft]{organization={Visioncraft OÜ},
            addressline={Maakri tn 19/1}, 
            city={Tallin},
            country={Estonia}}

\begin{abstract}
\revision{\textbf{Background} -- Prescriptive Process Monitoring systems aim at recommending, during the execution of a business process, interventions that, if followed, prevent poor performance of the process. Such interventions have to be i) reliable: they have to guarantee the achievement of the desired outcome or performance and ii) flexible: they cannot overturn the normal process execution.

\textbf{Problem} -- Most of the Prescriptive Process Monitoring solutions perform well in terms of recommendation reliability but provide the users with recommendations expressed in terms of specific activities that have to be executed without caring about their feasibility.

\textbf{Method} -- We propose a new Outcome-Oriented Prescriptive Process Monitoring system recommending temporal relations among activities that have to be guaranteed during the process execution. The proposed system is based on a Machine Learning model that learns the correlations between temporal relations among activities and the (positive) outcome of the process. Then, given the prefix of an ongoing process, the model is queried to return the most promising recommendations.

\textbf{Contribution} -- The main contribution is that the proposed system softens the mandatory execution of an activity at a given point in time, thus leaving more freedom to the user in deciding the interventions to put in place. This is achieved by providing recommendations that are expressed as Linear Temporal Logic formulas over activities.

\textbf{Results} -- The proposed system has been widely assessed using a pool of 22 real-life datasets. The results demonstrate the reliability of the provided recommendations by achieving an F1 score higher than 90\% on 18 datasets out of 22.}
\end{abstract}


\end{frontmatter}

\section{Introduction}
\revision{One of the main trends in Industry 4.0 refers to predictive monitoring and recommendations based on the data coming from information systems, machines, and IoT sensors. Predictive monitoring allows users to perform cost-effective interventions and determine them ahead of time before a (costly) system failure/process negative outcome occurs. On average, predictive monitoring (and maintenance) increases productivity by 25\%, reduces failures by 70\% and lowers costs by 25\%. \footnote{\url{https://www2.deloitte.com/content/dam/Deloitte/de/Documents/deloitte-analytics/Deloitte_Predictive-Maintenance_PositionPaper.pdf}} The focus of this paper is on Prescriptive Process Monitoring~\cite{TeinemaaTLDM18,DBLP:journals/kais/Fahrenkrog-Petersen22}, a field that provides a set of techniques that perfectly fit this Industry 4.0 trend.}

Prescriptive Process Monitoring is a branch of Process Mining~\cite{Aalst16book} that, leveraging historical process data recorded in an event log, aims at providing users with recommendations that, when followed during the execution of a business process, improve the probability of avoiding negative outcomes, or optimizing performance indicators. For example, a Prescriptive Process Monitoring system might recommend the interventions to carry on, or the activities to execute in order to minimize the likelihood of a patient going to intensive care, or the time required for dismissing a patient from a hospital.

The recommended interventions have to be \emph{reliable}, that is, they have to guarantee that the desired outcome or a good process performance is achieved, but, at the same time, they have to be \emph{flexible} enough to avoid recommending interventions that cannot be realized, for instance because a certain activity cannot be executed at a certain point in time during the process execution, or because it has a cost that cannot be afforded.
Most of the existing state-of-the-art approaches in the Prescriptive Process Monitoring field, however, mainly focus on returning reliable predictions, while neglecting the flexibility aspect. Most of them, indeed, do not take into account whether the recommended intervention is feasible or affordable in terms of costs. 

In this paper, we propose a new Outcome-Oriented Prescriptive Process Monitoring system that aims at ensuring not only the reliability of the provided recommendations, but also their flexibility. In particular, the proposed system returns different recommendations expressed \revision{in terms of temporal relations among activities \cite{PeSV07}} to be preserved in order to maximize the likelihood of achieving a desired outcome (e.g., avoiding a negative outcome), or to optimize a performance indicator of interest. The returned recommendations are prioritized based on their predicted impact on the process outcome. Each recommendation is composed of a temporal relation and a corresponding advice, i.e., ``it cannot be violated'' or ``it has to be satisfied''. Recommendations based on temporal relations among activities, \revision{differently from recommendations based on activities (or sequences of activities) to be executed (e.g.,~\cite{WeinzierlSZM20,LeoniDR20,BranchiFGM0R22}),} provide the user with more flexibility in choosing the interventions that best fit the current circumstances of an ongoing process execution. Furthermore, additional flexibility is provided by the system since users can choose among different prioritized recommendations.

\revision{The processes that most can benefit of the proposed system are the ones that are unpredictable, variable and work in changeable environments. Indeed, when the recommendations are based on temporal rules among process activities, the process participants can focus on few relevant constraints that the process should satisfy thus having the flexibility of adapting the process executions to the specific circumstances (if the recommendations are kept under-specified, few rules allow for multiple execution paths). For these reasons, the proposed Prescriptive Process Monitoring approach can be seamlessly applied in the context of healthcare processes, disaster handling processes and of all the so-called knowledge intensive processes \cite{DBLP:journals/jodsn/CiccioM015} that are characterized by a high degree of variability.}

{The approach proposed in this paper consists of two steps. In a first step, an encoding based on temporal relations among activities is used to encode the historical process data recorded in an event log. The encoded log is then used to train a Machine Learning (ML) classifier. In a second step, given an ongoing process case  $\sigma_k$, the classifier is inspected in order to extract temporal relations among activities (\emph{classification rules}) characterizing cases similar to the ongoing execution and leading to the desired outcome. Our assumption is indeed that temporal relations characterizing executions similar to the ongoing case $\sigma_k$ and leading to the desired outcome can convey effective recommendations towards that outcome. The proposed solution has been evaluated on a pool of 22 real-life event logs that have already been used as a benchmark in the Process Mining community \cite{TeinemaaDRM19}.
\revision{The experiments show that providing flexible recommendations does not affect their reliability. In addition, we also demonstrate the scalability of the approach by computing the execution times when the recommender system is applied to real-life logs. In the evaluation, we also provide a comparison in terms of different characteristics (recommendation target, input features, modeling techniques, type of recommendations) between the proposed system and  state-of-the-art Prescriptive Process Monitoring approaches.}

The paper is structured as follows: Section \ref{sec:background} introduces the main concepts useful for understanding the paper. Section~\ref{sec:method} and Section~\ref{sec:example} introduce the proposed approach and its application in a concrete use case. In Section~\ref{sec:evaluation}, a wide experimentation is presented, while Section~\ref{sec:conclusion} concludes the paper and spells out directions for future works.}

\revision{
\section{Problem Statement}
We define here the problem statement and clarify the research gap we want to fill using an example from the healthcare application domain. In particular, we consider a process for treating patients with hip fractures. If a patient is diagnosed with a hip fracture, the sequence of activities that is recommended as ``default'' procedure is: (1) bringing the patient into a room where pre-surgery anesthesia is delivered; (2) bringing the patient into an operating room where the surgery is performed; and (3) prescribing the patient with post-surgery physiotherapy for mobilizing the hip. 

In some cases, however, this strict sequence of activities is not possible based on the specific conditions of the patient. For example, surgical interventions are avoided or postponed whenever the patient suffers from other conditions that can cause additional complications. For example, if the patient is diagnosed with a chest infection and is treated with an amoxicillin therapy, the general anesthesia required for performing a surgery cannot be performed together with this therapy. Therefore, a more flexible recommendation to ``eventually'' perform the surgery allow the doctor to decide to postpone the surgery and perform it later on when the chest infection is solved. Similarly, the mobilization therapy should be delayed in case of leg pain after the hip surgery reported during a regular post-operational assessment. In this case, the mobilization cannot occur immediately after the surgery but the patient must be first prescribed with an analgesia therapy for pain relief.

The above example shows that there are situations (especially in unpredictable processes like healthcare treatments) in which recommending a specific sequence of activities is not a viable solution. However, the Prescriptive Process Monitoring systems available in the literature suffer from this rigidity. In the example provided, they would only be able to recommend the most common path that has to be executed if a patient is diagnosed with hip fracture, i.e.,  \textit{pre-surgery anesthesia -- surgery -- physiotherapy}, and do not allow the process participants to adapt the default modes of executing a process to specific contingencies. This is the research gap on which we build our contribution. In particular, in this paper, we show that it is possible to provide ``loose'' recommendations that are still strongly reliable. In particular, with respect to the existing approaches we:
\begin{enumerate}
  \item introduce an encoding of traces based on temporal relations among activities;
  \item use the encoding to build a new type of recommendations that ensure both reliability and flexibility;
  \item provide a mechanism to effectively prioritize the recommendations.
\end{enumerate}
Using the type of recommendations we propose, we can express suggestions like ``hip fracture diagnosis must be eventually followed by surgery'', ``amoxicillin therapy cannot coexist with pre-surgery anesthesia'', ``post-operational leg pain must be immediately followed by analgesia therapy'', and ``surgery must be eventually followed by physiotherapy''.
}
\section{Related Work}
\label{sec:rel_work}


Prescriptive Process Monitoring methods can be categorized according to whether the recommended interventions focus on the control flow, on the resources, or on other perspectives~\cite{Kubrak21Quovadis}.

The interventions involving the control flow usually prescribe a set of activities to perform next~\cite{WeinzierlSZM20,nakatumba2012meta,LeoniDR20,HeberHS15,DetroSPLLB20,BranchiFGM0R22,KotsiasKBTM22,abs-2303-03572}.
The next best activity can be prescribed in different domains and to different users, e.g., to employment companies to help customers in finding the most suitable job~\cite{LeoniDR20}, {to business analysts to improve the execution time, the customer satisfaction or the service quality of a} process~\cite{WeinzierlSZM20}, or to doctors in order to identify the most appropriate treatment based on the conditions of a patient~\cite{DetroSPLLB20}. \revision{Reinforcement Learning (RL) has recently gained popularity in recommending the next best activity(ies). In~\cite{BranchiFGM0R22}, an RL approach is used to prescribe the next best activity(ies) to only one of the actors involved in the process, e.g., the bank in a loan request handling process or the police department in a process related to the traffic management. In~\cite{KotsiasKBTM22} Prescriptive Process Monitoring is paired with Predictive Process Monitoring for recommending activities to achieve a desired outcome. The activities recommended in~\cite{abs-2303-03572} are computed with RL and causal inference in order to estimate the effect of a recommendation.}

Another group of prescriptions focuses on the resource perspective~\cite{WibisonoNBP15,SindhgattaGD16,Yaghoibi2017Cycle,abdulhameed2018resource}, e.g., which resource should perform the next activity. Also in this case, prescriptions can be applied to different domains. For example, in~\cite{WibisonoNBP15}, prescriptions {are related to} which police officer is best suited for the next task based on their predicted performance in a driving license application process. In~\cite{SindhgattaGD16}, recommendations on the repairs to carry out are provided to mechanics to guarantee that they complete their work within a predefined time. \revision{The approach presented in~\cite{Yaghoibi2017Cycle} recommends the most performing resource (i.e., workers in this case) to be assigned with a pending work item. In~\cite{abdulhameed2018resource}, the resources are recommended by optimizing the overall resource assignment of a process.}

Few works prescribe interventions regarding both control flow and resources~\cite{ShoushD21,NezhadB11,BarbaWV11,BozorgiTDRP21,YangDSZFXBM17,BOZORGI2023102198,Thomas2017Recommending}. For instance, in~\cite{ShoushD21}, an intervention to make an offer to a client together with the specific clerk that is the most suitable one to carry out the task {are prescribed}. In~\cite{NezhadB11}, the next activity and the specialist that should perform it are recommended to resolve open tickets in an IT service management process. \revision{In~\cite{YangDSZFXBM17,Thomas2017Recommending}, the recommended next activities and resources are extracted from a prototypical trace belonging to the same cluster as the input trace. The approaches presented in~\cite{BozorgiTDRP21,BOZORGI2023102198} focus on interventions that reduce the cycle time of a case. The former uses a Random Forest to predict the effects of triggering a recommendation for reducing the cycle time; the latter extends the former by considering the cycle time and also binary outcomes of a case and by using explainable Machine Learning techniques to select the recommendations.}

Finally, a last group of works focus on other types of interventions. For instance, in~\cite{TeinemaaTLDM18,DBLP:journals/kais/Fahrenkrog-Petersen22}, the authors propose a method in which a cost-model is used to control the creation of alerts in order to reduce the projected cost for a particular event log. In~\cite{metzger2019proactive}, the trade-off between the earliness and accuracy of the predictions for proactive process adaptation is discussed. The approach presented in~\cite{MetzgerKP20} uses online reinforcement learning to learn when to initiate proactive process adjustments based on forecasts and their run-time dependability. In~\cite{ShoushD21}, the authors tackle the problem of recommending interventions for avoiding an undesired outcome when a limited amount of resources is available. \revision{This work is extended in~\cite{ShoushD22} by also considering the effects of triggering a recommendation at a certain time.}

The approach proposed in our work focuses on control flow but, differently from existing works, it does not merely prescribe a sequence of activities to perform next but, rather, a set of temporal constraints that have to be satisfied. Temporal relations among activities provide more sophisticated and flexible recommendations (since they do not require the mandatory execution of a certain activity at a given point in time).
\section{Background}
\label{sec:background}
In this section, we introduce the main concepts needed for understanding the remainder of the paper.
\subsection{Events, Traces and Logs}
The main basic concept in Process Mining~\cite{Aalst16book}
is the \emph{event record} (or simply \emph{event}) that represents {the occurrence of} an activity 
in a business process. An event is associated with three mandatory attributes: the \emph{event class} (or \emph{activity name}) that states the name of the activity the event refers to, the \emph{timestamp} that specifies when the event occurred and the \emph{case id}, which is an identifier of the case of the business process {in which}
the event occurred. For example, a hospital might carry out procedures for the treatment of Sepsis, whose executions are logged in the hospital information system. Each treated Sepsis case is labeled with a case id and every event during this treatment (for example, the triage in the emergency room or the administering of particular antibiotics) is associated only with this case. In general, {each} 
event represents an activity occurred at a {certain point in}
time in a given case. In addition, events can have other attributes {related to}
the {data} payload: the so-called \emph{event-specific attributes}. In our Sepsis case example, an event attribute for the {activity related to the administration of particular antibiotics is the \attr{dosage} attribute. Finally, \emph{case attributes} refer to the whole case and are shared by all the events in the same case.}
In our Sepsis case example, case attributes are the \attr{age} and the \attr{sex} of a patient affected by Sepsis and the corresponding values will be the same for {each}
event in the case. The value of a case attribute does not change during the case {execution, i.e.,} it is static. {The event-specific attributes, instead, are dynamic, as they change their value based on the event.}

We now provide some formal definitions.
\begin{definition}[Event] An $event$ is a tuple $(a, c, time, (d_1, v_1), \ldots, (d_m, v_m))$ where $a \in \Sigma$ is the activity name, $c$ is the case id, $time$ is the timestamp and $(d_1, v_1) \ldots, (d_m, v_m)$ (with $m \geq 0$) are the event or case attributes and their values.
\end{definition}
We denote with $\Sigma$ the set of all the activity names and with $\mathbb{E}$ the universe of all events.
A \emph{case} is the sequence of events generated by a given {process execution}.
\begin{definition}[Case] A case is a non-empty sequence $\sigma = \langle e_1,\ldots,e_{n} \rangle$ of events such that $\forall i \in \{1,\ldots,n\}, e_i \in \mathbb{E}$ and $\forall i,j \in \{1,\ldots,n\} \; e_i{.}c = e_j{.}c$, that is, all events in the sequence refer to the same case.
\end{definition}
Consistently with the literature on Process Mining, many business process tasks {focus only on }
the activity names of a case. Therefore, it is customary to perform the projection of the activity names from a case to a \emph{trace}.
\begin{definition}[Trace] A trace of a case $\sigma = \langle e_1, e_2, \ldots, e_n \rangle$ is the sequence of the activity names in $\sigma$, $\langle e_1.a, e_2.a, \ldots, e_n.a \rangle$.
\end{definition}

We denote with $\mathbb{S}$ the universe of all possible traces and we use the symbol $\sigma$ for indicating both cases and traces when there is no risk of ambiguity. An \emph{event log} \Log is a set of \emph{complete cases} (i.e., the cases recording the execution of complete process execution). {For instance, in the Sepsis example, we can consider an event log containing two cases $\sigma_1$ and $\sigma_2$ (see \figurename~\ref{toyexample}).}
The activity name of the first event in case $\sigma_1$ is \act{ER registration}{; this event} occurred at \emph{11:15 AM} and it refers to case A. The first two attributes are static and {related to}
the age (27) and the sex of the patient (male){, respectively}. These have the same values for all the events in the case. The other attributes are event-specific and show that {\attr{amountPaid} is $10$ for the first event and $15$ for the last one.}
Note that not all events carry every possible event attribute. For example, the first event of case $\sigma_2$ does not have the attribute \attr{amountPaid}. 

\begin{figure*}
			\centering
\parbox{0.6\linewidth}{
            \footnotesize
			\begin{tabbing}
		$\sigma_1 =$ \=  $\langle (\act{ER registration}, A, 11:15AM, (\attr{age}, 27), (\attr{sex}, male), (\attr{amountPaid}, 10), (\attr{department}, centralDept)) \ldots$ \\ 
		                       \> $(\act{ER triage}, A, 11:35AM, (\attr{age}, 27), (\attr{sex}, male),  (\attr{amountPaid}, 15), (\attr{department}, NursingWard) \rangle$ \\
		$\sigma_2 = $ \= $ \langle (\act{order blood}, B, 3:20PM, (\attr{age}, 69), (\attr{sex}, female), (\attr{department}, GeneralLab) \ldots$ \\ 
		                       \> $(\act{payment}, B, 4:30PM, (\attr{age}, 69), (\attr{sex}, female), (\attr{amountPaid}, 100), (\attr{deparment}, FinancialDept) ) \rangle $
		                       \end{tabbing}}
		\caption{Extract of an event log.}
		\label{toyexample}
\end{figure*}

Given a case $\sigma =  \langle e_1, \ldots, e_n \rangle$ and a positive integer $k < n$, $\sigma_k = \langle e_1, \ldots, e_k \rangle$ is the \emph{prefix of $\sigma$} of length $k$. Furthermore, we define the \emph{prefix log} as the log composed of all possible case prefixes, 
which is typically used in Predictive and Prescriptive Process Monitoring settings~\cite{TeinemaaDRM19}.
\begin{definition}[Prefix Log]
Given a log \Log, the \emph{prefix log} $\Log^*$ of \Log is the event log that contains all prefixes of \Log, i.e., $\Log^* = \{\sigma_k : \sigma \in \Log, 1 \leq k < |\sigma| \}$.
\end{definition}

\subsection{\declare}
\label{sec:declare}
As stated above, the recommendations of the proposed Prescriptive Process Monitoring system are given in the form of temporal relations among activities to be performed during the execution of a process. Such recommendations have to be {expressed}
in a clear semantics for users. {To this aim, }
as a formal basis for specifying such temporal relations/patterns, we adopt the customary choice of Linear Temporal Logic over finite traces (\ltlf)~\cite{DeVa13}. This logic is at the basis of the well-known \declare~\cite{PeSV07} constraint-based process modeling language.

\ltlf has exactly the same syntax as standard \ltl, but, differently from \ltl, it interprets formulae over an unbounded, yet finite linear sequence of states. Given an alphabet $\Sigma$ of atomic propositions (in our setting, it represents the activity names of events), an \ltlf formula $\varphi$ is built by extending propositional logic with temporal operators:
\[\varphi ::= a \mid \lnot \varphi \mid \varphi_1\lor \varphi_2
 \mid \lnext\varphi \mid \lglobally\varphi \mid \lfuture\varphi \mid \varphi_1\luntil\varphi_2, \quad \text{ where $a \in \Sigma$.}\]

The semantics of \ltlf is given in terms of \emph{finite traces} denoting finite, \emph{possibly empty} sequences $\sigma=\tup{a_0, \ldots, a_n}$ of elements of $2^\Sigma$, containing all possible propositional interpretations of the propositional symbols in $\Sigma$. In this paper, consistently with the literature on Process Mining, we make the simplifying assumption that in each point of the sequence, one and only one element from $\Sigma$ holds. Under this assumption, $\sigma$ becomes a total sequence of activity name occurrences from $\Sigma$, matching the standard notion of trace. Table \ref{tab:ltl_sem} shows the semantics of the \ltlf operators.
\begin{table}[t!]
\caption{Semantics of the main \ltlf operators.}
\label{tab:ltl_sem}
\centering
\scalebox{0.8}{
\begin{tabular}{cl}
\toprule
\textbf{Operator} & \textbf{\ltlf Semantics} \\
\midrule
$\lnext\varphi$ & $\varphi$ has to hold in the next position of a sequence. \\
\midrule
$\lglobally\varphi$  & $\varphi$ has to hold always (Globally) in the subsequent positions of a sequence. \\
\midrule
$\lfuture\varphi$ &  $\varphi$ has to hold eventually (in the Future) in the subsequent position of a sequence. \\
\midrule
$\varphi\luntil\psi$ & $\varphi$ has to hold in a sequence at least Until $\psi$ holds. $\psi$ must hold in the current or in a future position.\\
\bottomrule
\end{tabular}
}
\end{table}

Given a trace $\sigma$, the evaluation of a formula $\varphi$ is done in a given position of the trace, and the notation $\sigma,i\models \varphi$ is used to express that $\varphi$ holds in position $i$ of $\sigma$. The notation $\sigma \models \varphi$ is used as a shortcut for $\sigma,0\models\varphi$, that is, to indicate that $\varphi$ holds over the entire trace $\sigma$ starting from the very beginning. A formula $\varphi$ is \emph{satisfiable} if it admits at least one trace $\sigma$ such that $\sigma \models \varphi$. A set of formulae $\mathcal{M} = \{\varphi_0, \dots, \varphi_n \}$ is a \emph{model} for a log $\Log$, denoted with $\Log \models \mathcal{M}$, if $\sigma \models \varphi_0 \wedge \ldots \wedge \varphi_n $ for each $\sigma \in \Log$. \revision{Checking whether $\Log \models \mathcal{M}$ holds is an important task in Process Mining and is called \textit{conformance checking}~\cite{PEEPERKORN2023106393}.}

\declare~\cite{PeSV07} is a declarative process modeling language based on \ltlf. More specifically, a \declare model fixes a set of activities, and a set of constraints over such activities, formalized using \ltlf formulae. The overall model is then formalized as the conjunction of the \ltlf formulae expressing its constraints. Among all possible \ltlf formulae, \declare selects some predefined patterns. Each pattern is represented as a \declare template, i.e., a formula with placeholders to be {replaced} 
by concrete activities to obtain a constraint. We denote placeholders in \declare\ templates with capital letters and concrete activities in \declare\ constraints with lower case letters. {Table~\ref{tab:timed-mfotl} reports the main \declare templates together with their \ltlf semantics and a textual description.} 

For binary constraints (i.e., constraints involving two activities), one of the two activities{, i.e., the activity triggering the constraint,} is called \textit{activation}, and the other {one, i.e., the one that satisfies the constraint, is called} \textit{target}. For example, for constraint \constr{response} (\act{a}, \act{b}),  \act{a} is an activation, {since}
the execution of \act{a} forces \act{b} to be executed eventually. Event \act{b} is, {instead,} 
the target{, since it guarantees the constraint satisfaction}. An activation of a constraint can be a \textit{fulfillment} {(if there is a target that satisfies the activation)}, or a \textit{violation} for such a constraint. When a trace satisfies a constraint, every activation of the constraint in the trace leads to a fulfillment. For example, {constraint} \constr{response} (\act{a}, \act{b}) is activated and fulfilled twice in trace $\langle \act{a}, \act{a}, \act{b}, \act{c} \rangle$, whereas, in trace $\langle \act{a}, \act{b}, \act{c}, \act{b} \rangle$, the same constraint is activated and fulfilled only once. When a trace does not satisfy a constraint, an activation of the constraint in the trace can lead to a fulfillment, but also to a violation (at least one activation leads to a violation). In $\langle \act{a}, \act{b}, \act{a}, \act{c} \rangle$, for example, {constraint \constr{response} (\act{a}, \act{b})} is activated twice and the first activation leads to a fulfillment (\act{b} occurs eventually), but the second activation leads to a violation (\act{b} does not occur {after the second activation}). A \textit{pending} activation is an activation that is not fulfilled in a prefix of a trace. For example, given the prefix $\langle \act{a}, \act{a}, \act{b}, \act{a}, \act{c} \rangle$ of a certain trace, constraint \constr{response} (\act{a}, \act{b}) has one pending activation in the last occurrence of \act{a} since it is not currently followed by any occurrences of \act{b}, but can be satisfied in the future considering that the prefix is (by definition) not complete. We denote by $|activations|$, $|fulfillments|$, $|violations|$ and $|pendings|$ the number of activations, fulfillments, violations and pending activations in {a} trace, respectively.

{When} testing a trace for satisfaction over one of the \declare constraints, the presence of an activation in the trace triggers the clause verification, requiring the (non-)execution of an event containing the target in the same trace. The notion of activation is related to the notion of \emph{vacuity detection} in model checking~\cite{Beer01efficientdetection,kupf:vacu03}. For example, in constraint \constr{response} (\act{a}, \act{b}), if \act{a} never occurs in a trace, then the constraint is ``vacuously'' satisfied, that is, satisfied without showing any form of interaction with the trace. 

\declare\ {templates} 
can be gathered into four main groups according to their semantics~\cite{Pesic2008} {(see the \declare\ {template} 
groups in Table~\ref{tab:timed-mfotl})}:
\begin{description}
\item[Existence $\mathcal{E}$:] the templates in this group {have only one parameter and check either the number of its occurrences in a trace or its position in the trace.}
\item[Choice $\mathcal{C}$:] the templates in this group {have two parameters and check if (at least) one of them occurs in a trace. 
} 
\item[Positive Relations $\mathcal{PR}$:] the templates in this group {have two parameters and check the relative position between the two corresponding activities.}
\item[Negative Relations $\mathcal{NR}$:] the templates in this group {have two parameters and check that the two corresponding activities do not occur together or do not occur in a certain order.}
\end{description}

{Hereafter, we denote with $\mathcal{A}$ the set of \declare templates, i.e., $\mathcal{A} = \mathcal{E} \cup \mathcal{C} \cup \mathcal{PR} \cup \mathcal{NR}$. Furthermore, given a set of activities $\Sigma$, we denote with $\mathcal{A}_\Sigma$ the set of \declare templates instantiated over activities in $\Sigma$.}
\begin{table}[h!t]
\caption{\ltlf semantics and textual description of the \declare templates. }
\label{tab:timed-mfotl}
\centering
\resizebox{\textwidth}{!}{
\begin{tabular}{lll}
\toprule
\textbf{Family: Template} & \textbf{\ltlf Semantics} & \textbf{Description}\\
\midrule
$\mathcal{E}$: \constr{existence} (n, \act{A}) & $\lfuture(\act{A} \wedge \lnext existence(n-1, \act{A}))$ & \act{A} has to occur at least $n$ times.\\
$\mathcal{E}$: \constr{absence} (n + 1, \act{A}) & $\neg existence(n+1, \act{A})$ & \act{A} has to occur at most $n$ times.\\
$\mathcal{E}$: \constr{exactly} (n, \act{A}) & $existence(n, \act{A}) \wedge absence(n+1, \act{A})$ & \act{A} has to occur exactly $n$ times.\\
$\mathcal{E}$: \constr{init} (\act{A}) & $\act{A}$ & Each case has to start with \act{A}.\\
\midrule
$\mathcal{C}$: \constr{choice} (\act{A}, \act{B}) & $\lfuture \act{A} \vee \lfuture \act{B}$ & \act{A} or \act{B} have to occur at least once.\\
$\mathcal{C}$: \constr{exclusive choice} (\act{A}, \act{B}) &$(\lfuture \act{A} \wedge \neg \lfuture \act{B}) \vee (\neg \lfuture \act{A} \wedge \lfuture \act{B})$& \makecell[l]{\act{A} or \act{B} have to occur at least once\\but not both.}\\
\midrule
$\mathcal{PR}$: \constr{responded existence} (\act{A}, \act{B})  & $\lfuture \act{A} \rightarrow \lfuture \act{B}$ & If \act{A} occurs, \act{B} must occur as well\\
$\mathcal{PR}$: \constr{response} (\act{A}, \act{B}) &  $\lglobally(\act{A} \rightarrow \lfuture \act{B})$ & If \act{A} occurs, \act{B} must eventually follow.\\
$\mathcal{PR}$: \constr{alternate response} (\act{A}, \act{B})  & $ \lglobally(\act{A} \rightarrow \lnext(\neg \act{A} \luntil \act{B}))$ & \makecell[l]{If \act{A} occurs, \act{B} must eventually follow\\without any other \act{A} in between.}\\
$\mathcal{PR}$: \constr{chain response}(\act{A}, \act{B}) &  $\lglobally(\act{A} \rightarrow \lnext \act{B})$ & If \act{A} occurs, \act{B} must occur next.\\
$\mathcal{PR}$: \constr{precedence}(\act{A}, \act{B}) &  $(\neg \act{B} \luntil \act{A}) \vee \lglobally(\neg \act{B})$ & \act{B} can occur only if \act{A} has occurred before.\\
$\mathcal{PR}$: \constr{alternate precedence (}\act{A}, \act{B}) & $(\neg \act{B} \luntil \act{A}) \wedge \lglobally( \act{B} \rightarrow \lnext((\neg \act{B} \luntil \act{A}) \vee \lglobally(\neg \act{B}))$) & \makecell[l]{\act{B} can occur only if \act{A} has occurred before,\\without any other \act{B} in between.}\\
$\mathcal{PR}$: \constr{chain precedence} (\act{A}, \act{B}) & $\lglobally(\lnext (\act{B}) \rightarrow \act{A})$ & \act{B} can occur only immediately after \act{A}.\\
\midrule
$\mathcal{NR}$: \constr{not responded existence} (\act{A}, \act{B}) & $\lfuture \act{A} \rightarrow \neg \lfuture \act{B}$ & If \act{A} occurs, \act{B} cannot occur. \\
$\mathcal{NR}$: \constr{not response} (\act{A}, \act{B}) & $\lglobally (\act{A} \rightarrow \neg(\lfuture \act{B}))$ & If \act{A} occurs, \act{B} cannot eventually follow. \\
$\mathcal{NR}$: \constr{not precedence} (\act{A}, \act{B}) & $\lglobally(\lfuture \act{B} \rightarrow \neg \act{A})$ & \act{A} cannot occur before \act{B}.\\
$\mathcal{NR}$: \constr{not chain response} (\act{A}, \act{B}) & $\lglobally(\act{A} \rightarrow \lnext(\neg \act{B}))$ & If \act{A} occurs \act{B} cannot occur next.\\
$\mathcal{NR}$: \constr{not chain precedence} (\act{A}, \act{B}) & $\lglobally(\lnext \act{B} \rightarrow \neg \act{A})$ & \act{A} cannot occur immediately before \act{B}.\\
\bottomrule
\end{tabular}
}
\end{table}

When a process execution is ongoing, the satisfaction of the corresponding trace prefix against a \declare\ constraint is not boolean. In particular, the \emph{Runtime Verification (RV) satisfaction value} of a \declare constraint $\varphi$ in a trace prefix $\sigma_i$ (indicated as $[\sigma_i\models \varphi]_{RV}$) is defined according to the four-valued semantics introduced in \cite{DBLP:conf/edoc/MaggiMB19}. In particular, a constraint in an ongoing process execution can be:
\begin{description}
\item[Possibly Satisfied:] the constraint is satisfied in the current position of the trace, but might be violated in the future.
\item[Possibly Violated:] 
{the} constraint is violated in the current position of the trace, but might be satisfied in the future.
\item[Satisfied:] the constraint is \emph{permanently satisfied} and can no longer become violated in the future positions of the trace.
\item[Violated:] the constraint is \emph{permanently violated} and can no longer become satisfied in the future positions of the trace.
\end{description}

The RV satisfaction value of a constraint in a trace depends on the type of constraint. Table~\ref{tab:semanticCriterion} shows the criteria to determine the RV satisfaction value of a constraint in a trace for each \declare\ template.
\begin{table}[h!t]
\caption{Criteria for identifying the RV satisfaction values of a constraint in a trace, where $a=|activations|$, $f=|fulfillments|$, $v=|violations|$, $p=|pendings|$, and $done$ is a boolean value specifying whether the trace is complete or not.}
	\label{tab:semanticCriterion}
	\centering
	\resizebox{\textwidth}{!}{
	\begin{tabular}{lcccc}
		\toprule
		\textbf{Template} & \textbf{Poss.viol}  & \textbf{Poss.sat}  & \textbf{Viol}  & \textbf{Sat}  \\
		\midrule
		\makecell[l]{ \constr{response} \\
         \constr{responded existence} } & $!done \land  (p > 0) $ & $ !done \land (p = 0) $  & $ done \land (p > 0) $ & $ done \land (p = 0) $ \\
		\midrule
		\makecell[l]{ \constr{not response} \\
			\constr{not chain response} \\
		    \constr{precedence} \\
			\constr{not precedence} \\
		    \constr{absence (n + 1)} \\
			\constr{chain precedence} \\
			\constr{not chain precedence} \\
			\constr{alternate precedence} }    & - &  $ !done \land (v = 0) $  &   $v > 0 $ & $ done \land (v = 0) $ \\
		\midrule
		\constr{init}   & -  & -  &  $ v > 0 $ &  $ f > 0 $ \\
		\midrule
		\constr{existence (n)}   & $ !done \land (a < n) $  & -  & $ done \land (a < n) $ & $ a >= n $ \\
		\midrule
		\constr{exactly (n)}
			& $ !done \land (a < n) $  & $ !done \land (a = n) $  & $ (a > n) \vee (done \land (a < n))$ & $done \land (a = n)$ \\
		\midrule
		\constr{not responded existence}  & -  &  $ !done \land (v = 0) $  &  $ v > 0 $ & $ done \land (v = 0) $ \\
		\midrule
		\makecell[l]{\constr{chain response} \\
\constr{alternate response}
}
			 & $ !done \land (v = 0) \land (p > 0) $   &  $ !done \land (v = 0) \land (p = 0) $  &  $ (v > 0) \vee (done \land (p > 0))$ & $ done \land (v = 0) \land (p = 0) $ \\
		\midrule
		\constr{choice}   & $!done \land (a = 0) $  & -  & $done \land (a = 0) $  & $a > 0 $  \\
		\midrule
		\constr{exclusive choice} & $!done \land (a = 0) $ & $!done \land (v = 0) \land (a > 0)$ &  $ (v > 0) \vee (done \land (a = 0)) $ &  $ done \land (v = 0) \land (a > 0)$ \\
		\bottomrule
	\end{tabular}
	}
\end{table}

\section{Method}
\label{sec:method}
\tikzstyle{block} = [draw, fill=blue!20, rectangle, minimum height=3em, minimum width=6em]

Our Outcome-Oriented Prescriptive Process Monitoring system focuses on prescribing interventions on the process control flow (i.e., on the activities to be executed) {in order to maximize the likelihood of achieving a certain outcome}. 
{Specifically,} our system prescribes to {users \textit{temporal relations among activities} that have to be preserved or violated in order to achieve a desired outcome.}
{For example, {to minimize the likelihood of a patient going to intensive care in a Sepsis case,}
the prescription (at a certain point {in time} of the case) {could be} that activity \act{Antibiotics treatment} should be immediately followed by \act{Leucocytes test}.}

These prescriptions need to be:
\begin{description}
  \item[D1:] \textbf{reliable}, {to ensure} the achievement of {the desired outcome;}
  {\item[D2:] \textbf{flexible}, so as to provide users with enough freedom in the application of the suggested recommendations}.
\end{description}
 To meet these desiderata, we propose a Prescriptive Process Monitoring system that:
\begin{enumerate}
  \item encodes the traces of a historical event log with temporal relations {among activities (\textbf{D2});}
  \item learns with an ML classifier {correlations between these temporal relations and case outcomes (\textbf{D1});}
  \item generates a {prioritized list} of prescriptions/recommendations $\mathcal{R}$ (i.e., {an ordered list of} temporal relations to satisfy or violate) for an ongoing case {(\textbf{D2})}.
\end{enumerate}
Given $\mathcal{A}_\Sigma$, the set of \declare constraints {instantiated over the set $\Sigma$ of the activities in a log}, we define a prescription/recommendation $r \in \mathcal{R}$ as a pair $\langle \varphi, 
c\rangle$, {where} 
$\varphi \in \mathcal{A}_\Sigma$ {is a temporal constraint} and 
{$c$ is a condition specifying whether the constraint $\varphi$ has to be satisfied or violated when the next activities of the process case are performed. These recommendations guide the user }
to achieve the desired outcome.
{More} details {on these 
conditions are provided} in Section~\ref{sec:rec_gen}.
{For instance, in the example above, }
$\mathcal{R} = \{\langle \varphi, 
c\rangle\}$ contains a unique recommendation $\langle \varphi, c\rangle$, where $\varphi$ is constraint \constr{chain response} (\act{Antibiotics treatment}, \act{Leucocytes test}) and 
$c$ is the condition ``{It} should not be violated''.

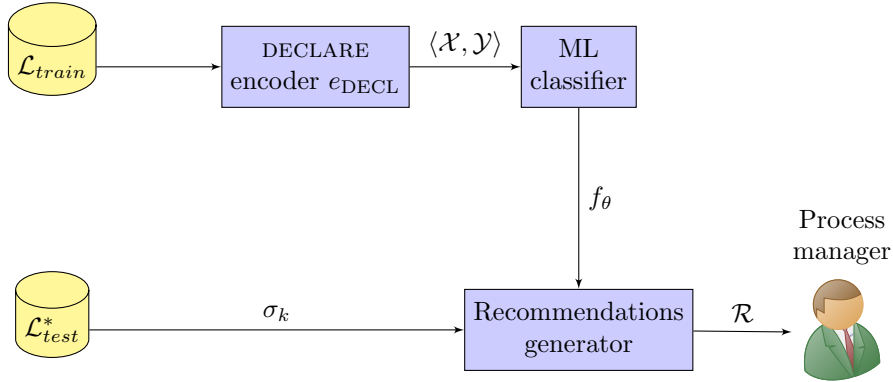
\begin{figure}[ht]
\centering
\tikzstyle{int}=[draw, fill=blue!20, minimum size=2em]
\begin{tikzpicture}[node distance=3.5cm, auto, >=latex', database/.style={
   cylinder, 
   cylinder uses custom fill, 
   cylinder body fill=yellow!50, 
   cylinder end fill=yellow!50, 
   shape border rotate=90,
   aspect=0.5,
   draw
  }]
   \node [database] (a) {$\Log_{train}$};
  \node [int] (b) [right of=a] {\makecell{$\declare$\\encoder  $e_{\mathrm{DECL}}$}};
  \node [int] (c) [right of=b] {\makecell{ML\\classifier}};
  \node [int] (d) [below of=c] {\makecell{Recommendations\\generator}};
  \node [database, below of=a] (test) {$\Log^*_{test}$};
  \node[businessman, minimum size=1.cm, right of=d, name=builder, label=above:\makecell{Process\\manager}](end) {};
  \path[->] (a) edge node {} (b);
  \path[->] (b) edge node {$\langle\mathcal{X},\mathcal{Y}\rangle$} (c);
  \draw[->] (c) edge node {$f_\theta$} (d) ;
  \draw[->] (test) edge node {$\sigma_k$} (d) ;
  \draw[->] (d) edge node {$\mathcal{R}$} (end) ;
\end{tikzpicture}
\caption{Overview of our Outcome-Oriented Prescriptive Process Monitoring system.}
\label{fig:system_overview}
\end{figure}

Figure~\ref{fig:system_overview} shows an overview of our proposal. Given a labeled training log $\Log_{train} = \{\langle \sigma_i, y(\sigma_i) \rangle \}_{i=1}^n$, in which each trace $\sigma_i$ is associated to a label $y(\sigma_i)$  (specifying whether a given desired outcome has been achieved or not in that trace), each trace $\sigma_i$ in $\Log_{train}$ is encoded, using a \declare\ encoder $e_{\mathrm{DECL}}$, in a feature vector $\bm{x}_i${. The feature vector is composed of $p$ features each representing the grounding on log activities of a \declare\ template, i.e., a \declare\ constraint $\varphi \in \mathcal{A}_\Sigma$} (Section~\ref{sec:encoding}). {Each trace is encoded based on whether the trace satisfies or not the \declare constraint associated to each feature.}
The encoded traces $\{\langle \bm{x}_i, y(\sigma_i) \rangle \}_{i=1}^n$ are then fed into an ML classifier to learn a classification task according to the given labeling (Section~\ref{sec:train}). The learned classifier $f_\theta$ is 
{then} queried by a generator of recommendations using the encoded prefixes of a prefix log $\Log^*_{test}$. The aim of the query is to extract from $f_\theta$ a set of recommendations $\mathcal{R}$ that maximize the likelihood of a positive outcome for a prefix $\sigma_k \in \mathcal{L}^*_{test}$ (Section~\ref{sec:rec_gen}).

\subsection{Encoding Traces Using \ltlf Temporal Patterns}
\label{sec:encoding}
{The proposed approach encodes temporal relations between log activities} 
by using a \emph{sequence encoder}. 
Each sequence $\sigma_i$ in $\Log_{train}$ or $\Log^*_{test}$ is transformed into a vector $e(\sigma_i)$.
\begin{definition}[Sequence/trace encoder] 
A sequence (or trace) encoder $e: \mathbb{S} \rightarrow \mathcal{X}_1 \times \cdots \times \mathcal{X}_p$ is a function that takes a (partial) trace $\sigma_i$ and transforms it into a feature vector $\bm{x_i} = e(\sigma_i)$ in the $p$-dimensional vector space $\mathcal{X}_1 \times \cdots \times \mathcal{X}_p$, with $\mathcal{X}_j \subseteq \mathbb{R}, 1 \leq j \leq p$ being the domain of the $j$-th feature.
\end{definition}
%
Specifically, we adopt an encoding based on the \declare semantics reported in Table~\ref{tab:semanticCriterion} so as to obtain, for each (prefix) trace $\sigma_i$, a feature vector $\bm{x}_i$. The features used to build the feature vector are obtained by instantiating all the \declare\ templates in Table \ref{tab:timed-mfotl} with all the combinations\footnote{The combinations do not consider repetitions since some of the (binary) \declare\ templates generate constraints that can never be satisfied (on finite traces) when activation and target correspond to the same activity.} of activities available in $\Log_{train} \cup \Log^*_{test}$ (representing the alphabet $\Sigma$). Each element of the feature vector is then a {value representing whether each of those \declare\ constraints $\varphi_j \in \mathcal{A}_\Sigma$ is (possibly) satisfied or (possibly) violated in $\sigma_i$.}

The possible feature values for the $j$-th feature are:
\begin{itemize}
\item 0, if the \declare constraint $\varphi_j$ is violated in $\sigma_i$;
\item 1, if the \declare constraint $\varphi_j$ is satisfied in $\sigma_i$;
\item 2, if the \declare constraint $\varphi_j$ is possibly violated in $\sigma_i$;
\item 3, if the \declare constraint $\varphi_j$ is possibly satisfied in $\sigma_i$.
\end{itemize}
After the encoding phase, the event log ($\Log_{train}$ or $\Log^*_{test}$) is transformed into a matrix of numerical values, where each row corresponds to a sequence and each column corresponds to a \declare constraint. Each entry is the  RV satisfaction value of the constraint in the sequence (expressed using an integer value as explained above). 

We present now an example of trace encoding, using trace $\sigma_i = \left\langle \act{a}, \act{b}, \act{c}, \act{a}, \act{b}, \act{c}, \act{c}, \act{a}, \act{b}\right\rangle${, alphabet $\Sigma = \{\act{a}, \act{b}, \act{c}\}$, and \constr{response} as the only template used in the encoding, so that we have $\mathcal{A}_\Sigma = \{$\constr{response} (\act{a}, \act{b}),  \constr{response} (\act{b}, \act{a}), \constr{response} (\act{a}, \act{c}),  \constr{response} (\act{c}, \act{a}), \constr{response} (\act{b}, \act{c}),  \constr{response} (\act{c}, \act{b})$\}$.} The feature values are determined using the semantics reported in Table~\ref{tab:semanticCriterion}. For example:
\begin{itemize}
\item if $\sigma_i$ is a complete trace, constraint \constr{response} (\act{a}, \act{c}) is violated since the third activation (the third occurrence of \act{a}) leads to a violation ({it} is not eventually followed by \act{c}).
{It} is {hence} encoded with 0;
\item if $\sigma_i$ is a complete trace, constraint \constr{response} (\act{a}, \act{b})  is satisfied.
{It} is {hence} encoded with 1;
\item if $\sigma_i$ is a prefix, constraint \constr{response} (\act{a}, \act{b}) is possibly satisfied since there are no pending activations for this constraint (but the constraint can still be violated in the future). The constraint is then encoded with 3;
\item if $\sigma_i$ is a prefix, constraint \constr{response} (\act{b}, \act{c}) is possibly violated since the last occurrence of \act{b} is a pending activation for this constraint. The constraint is {hence} encoded with 2.
\end{itemize}
{Assuming that the features used by the encoder are in the order 
$\langle$\constr{response} (\act{a}, \act{b}), \constr{response} (\act{b}, \act{a}), \constr{response} (\act{a}, \act{c}), \constr{response} (\act{c}, \act{a}), \constr{response} (\act{b}, \act{c}), \constr{response} (\act{c}, \act{b})$\rangle$, the corresponding feature vector is} $\bm{x}_i = \langle 1,0,0,1,0,1\rangle$ if $\sigma_i$ is a complete case, or $\bm{x}_i = \langle 3,2,2,3,2,3\rangle$ if $\sigma_i$ is a prefix. This operation if performed by the \declare encoder $e_{\mathrm{DECL}}$ in Figure~\ref{fig:system_overview}{, leveraging the semantics reported in Table~\ref{tab:semanticCriterion}, } where the parameter $done$ is $True$ when the training log $\Log_{train}$ is provided as input, $False$ when the test prefix log $\Log^*_{test}$ is provided as input. \revision{It is worth noticing that a missing activity in a trace is treated as a vacuous satisfaction and encoded with a proper numeric value. Therefore, there are no missing values in the encoded data.}

Differently from the standard methods for trace encoding~\cite{TeinemaaDRM19}, this kind of encoding allows {for} the generation of easily understandable recommendations $\mathcal{R}$ based on the simple and intuitive \declare temporal patterns. These {recommendations} will be extracted from an ML model $f_\theta$, trained with the event log $\Log_{train}$ encoded with the \declare\ encoding just introduced, by using a rule extraction technique. However, this kind of encoding has the drawback of creating very long feature vectors $\bm{x}_i$. Indeed, for {a set of activities $\Sigma$}
and a given \declare constraint involving an activation and a target {activity}, the number of generated features is $O(|\Sigma|^2)$. For instance, in a very simple domain where $|\Sigma| = 10$, the number of features generated using the templates shown in Table~\ref{tab:timed-mfotl} is $4*10 + 14*10^2 = 1440$, where 4 is the number of unary constraints and 14 the number of binary constraints. {These} large feature vectors have the disadvantage of including irrelevant or redundant features, {which} are time demanding for the ML algorithm used to train the classifier, {and make} the classification problem harder. We addressed this problem by adopting a feature selection strategy. In particular, using the Apriori algorithm described in \cite{agrawalApriori}, we select the most frequent (pairs of) activities according to a user-defined threshold and we instantiate the \declare\ templates only by using those activities. The obtained features are then ranked according to their mutual information score~\cite{ross2014mutual} with the class label. In our experiments, we set the Apriori algorithm threshold to 5\% in order to select a sufficiently high number of features to be ranked based on the mutual information score. The number of the top most informative features, instead, was selected through a grid search, as explained in Section~\ref{sec:setup}.


\subsection{Training a Classifier for Reliable Recommendations}
\label{sec:train}
Desideratum \textbf{D1} requires reliable recommendations, that is, recommendations that, if followed, help achieving {the desired process outcome. }
We therefore need an effective mapping function $f_\theta$ between the recommendations (expressed in terms of satisfaction or violation of certain \declare\ constraints) and the outcome of a sequence $\sigma_i$. The following definition formalizes the outcome of a complete trace with a known \emph{class label} given the set $\mathbb{S}$ of all possible sequences.
\begin{definition}[Labeling function]\label{def:labeling} A labeling function $y: \mathbb{S} \rightarrow \mathcal{Y}$ maps a trace $\sigma_i$ to its class label $y(\sigma_i) \in \mathcal{Y}$ with $\mathcal{Y}$ being the domain of the class labels.
\end{definition}
For classification tasks, $\mathcal{Y}$ is a finite set of categorical outcomes. In this paper, we only consider binary outcomes, i.e., $\mathcal{Y} = \{0,1\}$. {For instance,} in the Sepsis case example, a case $\sigma_i$ can be 
labeled  {as positive} ($y(\sigma_i)=1$) {if the patient does not need to go to intensive care,}
or {as negative} ($y(\sigma_i)=0$) in the opposite case. For building the mapping function $f_\theta$ between the feature vectors $\bm{x}_i = e_{\mathrm{DECL}}(\sigma_i)$ and their labels $y(\sigma_i)$, we train an ML classifier.
\begin{definition}[Classifier] A classifier $f_\theta: \mathcal{X}_1 \times \cdots \times \mathcal{X}_p \rightarrow \mathcal{Y}$ is a function that takes an encoded vector $\bm{x} \in \mathcal{X}_1 \times \cdots \times \mathcal{X}_p$ and estimates its class label.
\end{definition}
The variable $\theta$ is characteristic of the classifier and indicates a set of parameters to be learned to have a reliable estimation of the class label. This set of parameters is learned by training the classifier through a learning algorithm whose input is the training log $\Log_{train}$. 

\subsection{Generating Recommendations as \ltlf Temporal Patterns}
\label{sec:rec_gen}
The trained ML classifier $f_\theta$ is successively used to extract a {prioritized list}
of recommendations $\mathcal{R}$ for a given prefix $\sigma_k \in \Log^*_{test}$. As ML classifiers, we {use Decision Trees (DTs)}
as (i) {they have shown good performance in Predictive Process Monitoring~\cite{TeinemaaDRM19} (\textbf{D1}); (ii) the most important features are explicitly available in the model and recommendations can be sorted based on their discriminativeness (\textbf{D2}). In particular, in DTs, the features closest to the root are the most discriminative ones. This provides a natural prioritization on the effectiveness of the recommendations.}
\revision{The process of building a decision tree DT from a log with the \declare encoding is formalized in Algorithm \ref{algo:DecisionTreeInduction}.}
\begin{algorithm}[h!tbp]
\caption{DecisionTreeInduction}
\label{algo:DecisionTreeInduction}
\textbf{Input}: $\Log, \mathcal{A}_\Sigma$\\
\textbf{Output}: $DT$\\
\begin{algorithmic}[1] 
\STATE $\langle \mathcal{X}, \mathcal{Y} \rangle = e_{\mathrm{DECL}}(\Log, \Sigma_k)$ \COMMENT{\declare encoding of the input log}
\STATE $DT = \mathrm{DTInduction}(\mathcal{X}, \mathcal{Y})$ \COMMENT{Construction of the Decision Tree}
\STATE \textbf{return} $DT$
\end{algorithmic}
\end{algorithm}

A path from the root to a leaf of a DT 
simply consists of a set of features along with their corresponding values learned during the training process. Each trace is mapped to a path in the DT that can be used to classify it. The paths in the DT are identified using decision points expressed as conditions on the feature values. Therefore, the path itself can be seen as a classification rule for an input trace.
In our case, a classification rule for a (complete) trace $\sigma_i$ has the following form: \texttt{IF} $([\sigma_i\models \varphi_0]_{RV} = val_0) \wedge , \ldots,  \wedge$ $([\sigma_i\models \varphi_n]_{RV} = val_n) \mbox{ \texttt{THEN} } y(\sigma_i)$, where $\varphi_j \in \mathcal{A}_\Sigma$, and, since the classifier is trained over complete traces, $val_j$ can be \emph{satisfied} if $\sigma_i \models \varphi_j$, or \emph{violated} otherwise. 
A good DT contains a set of paths that are able to discriminate, in an effective way, between $\Log_{train}^+ = \{\sigma_i \in \Log_{train}: y(\sigma) = 1 \}$ (the subset of traces of $\Log_{train}$ with a positive label) and $\Log_{train}^- = \{\sigma_i \in \Log_{train}: y(\sigma) = 0 \}$ (the subset of traces of $\Log_{train}$ with a negative label). Note that $\Log_{train}^+$ and $\Log_{train}^-$ represent a partition of $\Log_{train}$, i.e., $\Log_{train} = \Log_{train}^+ \cup \Log_{train}^-$ and $\Log_{train}^+ \cap \Log_{train}^-= \emptyset$. 

Before explaining the details of our method to generate recommendations, we introduce some preliminary notions. Given a DT, let $\mathcal{P} = \{p_0, p_1, \ldots \}$ be the set of its paths from the root to the leaves. A single path $p$ from the root to a leaf is defined as:
$$
\langle (\varphi_0, val_0), (\varphi_1, val_1),  \ldots , polarity, impurity, \#PosSamples, \#NegSamples\rangle,
$$
where $(\varphi_0, val_0), (\varphi_1, val_1),  \ldots $ are the feature-value pairs belonging to the path, \emph{polarity} and \emph{impurity} are the majority class and the impurity value (computed by using either the Gini index or the entropy) of the leaf node of the path, and \emph{\#PosSamples} and \emph{\#NegSample} are the number of positive and negative training samples matching the path.

Given a trace prefix $\sigma_k$, our proposal is to derive a set of recommendations $\mathcal{R}$ from $\mathcal{P}$ by finding a positive path $p \in \mathcal{P}^+$ (that is, a path with a positive polarity {and hence likely to lead to a positive outcome}) with feature-value pairs {matching as much as possible the ones appearing in the encoding of} $\sigma_k$. Our assumption is that a positive path very similar {(according to a similarity score)} to $\sigma_k$ {can convey} 
effective recommendations for {achieving} a positive {process} outcome. 
However, the similarity between a path and a prefix {could be not sufficient to find a unique path providing good recommendations.} %
Indeed, for short prefixes, 
many paths in $\mathcal{P}$ could have the same similarity score as $\sigma_k$, due to the small number of activities in $\sigma_k$. To better discriminate {among the different paths with the same similarity score,} 
we therefore select the path with {the} lowest impurity and {the} highest probability. We formalize these ideas with the notion of \emph{recommendation score} $\rho(\sigma_k, p)$ between a prefix $\sigma_k$ and a path $p$, defined as:
\begin{equation}
\label{eq:rho}
\rho(\sigma_k, p) = \lambda_1\mathcal{F}(\sigma_k, p) + \lambda_2(1-impurity(p)) + \lambda_3\frac{\#PosSamples(p)}{\sum_{p_j \in \mathcal{P}^+}\#PosSamples(p_j)},
\end{equation}
where $\mathcal{F}$ is a \emph{fitness function} {measuring the similarity between $\sigma_k$ and path $p$, }the term weighted by $\lambda_2$ {refers to}
the \emph{purity} of the leaf node of $p$ (i.e., the complement of its impurity) and the term weighted by $\lambda_3$ is the probability of path $p$ classifying correctly a positive sample ($\mathcal{P}^+$ is the set of paths {leading to a positive outcome). All the weighted terms of Eq.\ (\ref{eq:rho}) are numbers between 0 and 1, and weights $\lambda_1$, $\lambda_2$ and $\lambda_3$ are hyperparameters of the generation algorithm such that $\lambda_1 + \lambda_2 + \lambda_3 = 1$.}
{The fitness function  $\mathcal{F}$ is computed as}
the average {\emph{compliance}} of the learned satisfaction values of the \declare constraints in path $p$ and the RV satisfaction values of these constraints in prefix $\sigma_k$.
Given a path $p$, let $rule(p) = \langle(\varphi_0, val_0), (\varphi_1, val_1), \ldots \rangle$ be the sequence of pairs of constraints and their satisfaction values in path $p$. The fitness function $\mathcal{F}$ is then defined as:
\begin{equation}
\label{eq:fitness}
\mathcal{F}(\sigma_k, p) = \frac{1}{|rule(p)|}\sum_{(\varphi, val) \in rule(p)}\mathcal{C}(val, [\sigma_k\models \varphi]_{RV}),
\end{equation}
where the {compliance}
function $\mathcal{C}$ returns higher values if the learned satisfaction value for $\varphi$ is similar to the RV satisfaction value of $\varphi$ in $\sigma_k$, $[\sigma_k\models \varphi]_{RV}$. {Specifically}:
\begin{equation}
\label{eq:compliance}
\mathcal{C}(val, [\sigma_k\models \varphi]_{RV})= \begin{cases}
1 &\text{if $val = $ {violated}, and $[\sigma_k\models \varphi]_{RV} = $ {violated/possibly violated}}\\
1 &\text{if $val = $ {satisfied}, and $[\sigma_k\models \varphi]_{RV} = $ {satisfied/possibly satisfied}}\\
0.5 &\text{if $val = $ {violated}, and $[\sigma_k\models \varphi]_{RV} = $ {possibly satisfied}}\\
0.5 &\text{if $val = $ {satisfied}, and $[\sigma_k\models \varphi]_{RV} = ${possibly violated}}\\
0 &\text{otherwise}.
\end{cases}
\end{equation}


Given a prefix $\sigma_k$ and a DT with $\mathcal{P}^+$ the set of paths leading to a positive outcome in DT, we define the path $p^*$ that conveys the best recommendations as the positive path that maximizes the recommendation score $\rho$ with $\sigma_k$:
\begin{equation}
\label{eq:best_path}
p^* = \argmax_{p \in \mathcal{P}^+}\rho(\sigma_k, p).
\end{equation}

The extraction of the recommendations $\mathcal{R}$ from $p^*$ is straightforward. Let $rule(p^*)$ be the set of constraints and their satisfaction values encoded in $p^*$. The recommendation is generated for each pair {$(\varphi, val)$} 
in $rule(p^*)$ by comparing again $val$ with $[\sigma_k\models \varphi]_{RV}$. 

\begin{table}[h!t]
\caption{Recommendation generation from the constraints in path $p^*$ and prefix $\sigma_k$.} 
\label{tab:rec_rules}
\centering
\scalebox{0.8}{
\begin{tabular}{cccc}
\toprule
$val$                      & $[\sigma_k\models \varphi]_{RV}$                  & {Rationale}                                              & Recommendation                     \\
\midrule
\texttt{Satisfied} &  \texttt{Violated}           & \makecell{{The case cannot be recovered anymore}}      & - \\
\texttt{Satisfied} &  \texttt{Satisfied}          & \makecell{{No action needs to be taken}}       & - \\
\texttt{Satisfied} &  \texttt{Possibly Violated}  & \makecell{{An action should be taken}}   & $\langle \varphi, \mbox{{It} should become satisfied} \rangle$ \\
\texttt{Satisfied} &  \texttt{Possibly Satisfied} & \makecell{{An action should be taken}} & $\langle \varphi, \mbox{{It} should not be violated} \rangle$ \\
\texttt{Violated}  &  \texttt{Satisfied}          & \makecell{{The case cannot be recovered anymore}}       & - \\
\texttt{Violated}  &  \texttt{Violated}           & \makecell{{No action needs to be taken}}       & - \\
\texttt{Violated}  &  \texttt{Possibly Violated}  & \makecell{{An action should be taken}} & $\langle \varphi, \mbox{{It} should not be satisfied} \rangle$ \\
\texttt{Violated}  &  \texttt{Possibly Satisfied} & \makecell{{An action should be taken}}  & $\langle \varphi, \mbox{{It} should become violated} \rangle$ \\
\bottomrule
\end{tabular}
}
\end{table}

Table~\ref{tab:rec_rules} shows the rules for generating recommendations from $val$ and $[\sigma_k\models \varphi]_{RV}$. The idea is to provide a {recommendation so that the ongoing trace $\sigma_k$ becomes compliant with the classification rule. A full compliance between $val$ and $[\sigma_k\models \varphi]_{RV}$ results in a case completely in line with the positive classification rule of the DT. In this case, no prescription is needed (see rows 2 and 6). On the other hand, if a contradiction occurs between $val$ and $[\sigma_k\models \varphi]_{RV}$, the case cannot be fixed anymore (see rows 1 and 5). In other situations, a recommendation $\langle \varphi, c\rangle$ is provided (see rows 3-4 and 7-8). The recommendation suggests the action to take and is composed of a \declare constraint $\varphi$ and a condition $c$ expressing whether the constraint should be satisfied or not.} 
\revision{Algorithm \ref{algo:RecommendationGeneration} summarizes the generation of recommendations from a log $\Log$ and an input prefix trace $\sigma_k$ of length $k$.} 
\begin{algorithm}[h!tbp]
\caption{RecommendationGeneration}
\label{algo:RecommendationGeneration}
\textbf{Input}: $DT, \sigma_k$\\
\textbf{Output}: $\mathcal{R}$\\
\begin{algorithmic}[1] 
\STATE $\mathcal{P}^+ = \mathrm{PositivePaths}(DT)$ \COMMENT{Extraction of the positive paths}
\STATE $p^* = \argmax_{p \in \mathcal{P}^+}\rho(\sigma_k, p)$ \COMMENT{Selection of the path with the  highest $\rho$, see Eq. \ref{eq:rho}}
\STATE $\mathcal{R} := \{ \}$
\FORALL{$\langle\varphi, val\rangle \in rule(p^*)$}
\IF {$val = \mathtt{Satisfied} $ and $[\sigma_k\models \varphi]_{RV} = \mathtt{Possibly Violated}$}
\STATE $\mathcal{R} := \mathcal{R} \cup  \{\langle \varphi, \textrm{It should become satisfied}\rangle\}$
\ELSIF{$val = \mathtt{Satisfied}$ and $[\sigma_k\models \varphi]_{RV} = \mathtt{Possibly Satisfied}$}
\STATE $\mathcal{R} := \mathcal{R} \cup  \{\langle \varphi, \textrm{It should not be violated}\rangle\}$
\ELSIF{$val = \mathtt{Violated}$ and $[\sigma_k\models \varphi]_{RV} = \mathtt{Possibly Violated}$}
\STATE $\mathcal{R} := \mathcal{R} \cup  \{\langle \varphi, \textrm{It should not be satisfied}\rangle\}$
\ELSIF{$val = \mathtt{Violated}$ and $[\sigma_k\models \varphi]_{RV} = \mathtt{Possibly Satisfied}$}
\STATE $\mathcal{R} := \mathcal{R} \cup  \{\langle \varphi, \textrm{It should become violated}\rangle\}$
\ENDIF
\ENDFOR

\STATE \textbf{return} $\mathcal{R}$
\end{algorithmic}
\end{algorithm}
\secRevision{Figure~\ref{fig:rec_generation} expands Figure~\ref{fig:system_overview} by showing a graphical overview of the recommendations generator step, that is, Algorithm~\ref{algo:RecommendationGeneration}.
\begin{figure}[ht]
\centering
\tikzstyle{int}=[draw, fill=blue!20, minimum size=2em]
\tikzstyle{input} = [coordinate]
\tikzstyle{sum} = [draw, fill=white, circle]
\begin{tikzpicture}[node distance=3.7cm, auto, >=latex', database/.style={
   cylinder, 
   cylinder uses custom fill, 
   cylinder body fill=yellow!50, 
   cylinder end fill=yellow!50, 
   shape border rotate=90,
   aspect=0.5,
   draw
  }]
   \node [input, name=input1] (a) {};
  \node [int] (b) [below of=a] {\makecell{Positive\\paths\\extraction}};
  \node [int] (c) [right of=b] {\makecell{$\rho$ computation\\and best\\path selection}};
  
  \node [int] (d) [right of=c] {\makecell{Recommendation\\composition\\with\\$[\sigma_k\models \varphi]_{RV}$}};
  \node [database, below of=b] (test) {$\Log^*_{test}$};
  \node [sum, right of=test] (sum) {};
  \node[businessman, minimum size=1.cm, right of=d, name=builder, label=above:\makecell{Process\\manager}](end) {};
  \path[->] (a) edge node {$f_\theta = DT$} (b);
  \path[->] (b) edge node {$\mathcal{P}^+$} (c);
  \draw[->] (c) edge node {$p^*$} (d) ;
  \draw[->] (test) edge node {$\sigma_k$} (sum) ;
  \draw[->] (d) edge node {$\mathcal{R}$} (end) ;
  \draw[->] (sum) edge node {} (c) ;
  \draw[->] (sum) -| node {} (d) ;
\end{tikzpicture}
\caption{Overview of the recommendation generation step.}
\label{fig:rec_generation}
\end{figure}
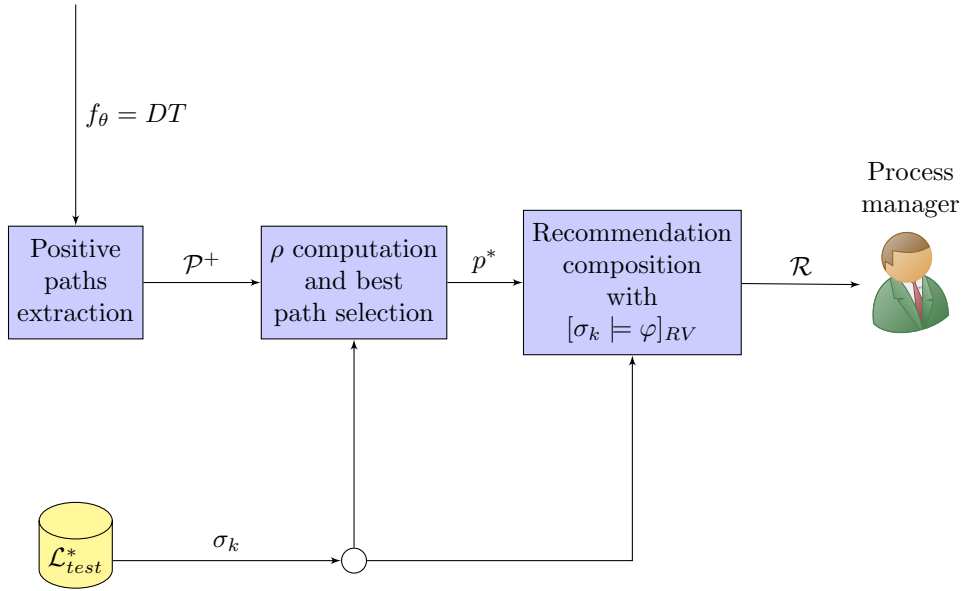
Line 1 of Algorithm~\ref{algo:RecommendationGeneration} performs the extraction of the positive paths (first block of Figure~\ref{fig:rec_generation}) from the DT returned by Algorithm~\ref{algo:DecisionTreeInduction}, that is, the parametric function $f_\theta$ in Figure~\ref{fig:system_overview}. Line 2 of Algorithm~\ref{algo:RecommendationGeneration} selects the best path $p^*$ according to the $\rho$ score between the prefix $\sigma_k$ and each path in $\mathcal{P}^+$ (second block of Figure~\ref{fig:rec_generation}). The best path $p^*$, together with the prefix $\sigma_k$, is then used to compose the recommendations in $\mathcal{R}$ (lines 4-13 and third block of Figure~\ref{fig:rec_generation}).
}

%
{The resulting list of recommendations $\mathcal{R}$ is returned already ordered by importance,} 
considering that the recommendations extracted from the feature-value pairs that, in the DT, are the closest to the root best discriminate between a positive and a negative outcome for $\sigma_k$. Therefore, the corresponding recommendations can be presented to the {users} 
with a {higher} priority to allow them to choose the most important recommendations to adopt in case it is not possible to follow all of them.

\secRevision{We now discuss the computational complexity of our proposal. Algorithm~\ref{algo:DecisionTreeInduction} is performed once to induce the decision tree from the log $\Log$ and its computational complexity is the sum of the computational complexity of the \declare encoding steps (line 1) and the one of the construction of the tree (line 2). The \declare encoding requires $O(|\Log|\times\mathcal{A}_\Sigma\times N)$ steps with $N$ being the length of the longest trace $\sigma$ in $\Log$. The decision tree construction requires $O(|\Log|\times\mathcal{A}_\Sigma\times log(|\Log|))$ steps. Algorithm~\ref{algo:RecommendationGeneration} is called for every prefix $\sigma_k \in \Log^*$ and it requires $O(k\times w \times |\mathcal{P^+}|)$ steps, where $k$ is the length of $\sigma_k$ and $w$ is the length of the longest path in $|\mathcal{P^+}|$. Here, the computation is dominated by the computation of the best path $p^*$ at line 2 that requires the checking of $w$ constraints over $\sigma_k$. Each constraint check is linear in the length $k$ of $\sigma_k$. Figure~\ref{fig:rec_times} reports the computational times for Algorithm~\ref{algo:RecommendationGeneration} that confirm this analysis.}

\section{An End-to-End Example for a Sepsis Treatment Use Case}
\label{sec:example}
We present now an end-to-end example of how our system works in a real-life Sepsis treatment use case. We consider one of the datasets used in our evaluation ($sepsis\_case\_2$) and the existence family of \declare templates $\mathcal{E}$. We choose this family as it allows us to provide a simple but concrete description of our Prescriptive Process Monitoring system. As stated in Section~\ref{sec:datasets}, the $sepsis\_case\_2$ dataset contains 782 cases. Among them, the cases with a positive label are the ones related to patients that do not need to go to intensive care. Therefore, our Prescriptive Process Monitoring system will provide recommendations for avoiding the admission of a patient to the intensive care.

\subsection{Preprocessing and encoding}
The dataset contains 24 activity names representing standard activities being performed during a Sepsis case. During a preprocessing phase, the system discards the activity names that are infrequent in the dataset, based on a user-defined threshold. In our case, we select the activity names that appear in at least 5\% of the cases in the dataset. After this step, 12 activity names remain: \act{ER Registration}, \act{ER Triage}, \act{ER Sepsis Triage}, \act{CRP}, \act{LacticAcid}, \act{Leucocytes}, \act{IV Liquid}, \act{IV Antibiotics}, \act{Admission NC}, \act{Release A}, \act{Return ER}, \act{Release B}. These activity names are combined with the existence templates in $\mathcal{E}$ (Table~\ref{tab:timed-mfotl}), thus obtaining 48 features for the trace encoding.

Assuming that our feature selection algorithm considers the top $h$ features, with $h$ corresponding to half of the number of the original features, after the feature selection, the resulting features are reduced to 24. The obtained features are in our case: \constr{existence} (\act{ER Registration}), \constr{existence} (\act{ER Triage}), \constr{existence} (\act{ER Sepsis Triage}), \constr{init} (\act{CRP}), \constr{exactly} (\act{CRP}), \constr{absence} (\act{LacticAcid}), \constr{absence} (\act{Leucocytes}), \constr{exactly} (\act{Leucocytes}), \constr{exactly} (\act{IV Liquid}), \constr{existence} (\act{IV Antibiotics}), \constr{existence} (\act{Admission NC}), \constr{absence} (\act{Admission NC}), \constr{exactly} (\act{Admission NC}), \constr{existence} (\act{Release A}), \constr{absence} (\act{Release A}), \constr{init} (\act{Release A}), \constr{exactly} (\act{Release A}), \constr{existence} (\act{Return ER}), \constr{absence} (\act{Return ER}), \constr{init} (\act{Return ER}), \constr{exactly} (\act{Return ER}), \constr{existence} (\act{Release B}), \constr{absence} (\act{Release B}), \constr{exactly} (\act{Release B}).
The dataset has 782 traces that are divided into a training set $\Log_{train}$ (625 traces, 80\%) and a test set (157 traces, 20\%) from which the prefix log $\Log^*_{test}$ is extracted. The traces in $\Log_{train}$ and $\Log^*_{test}$ are all encoded using the above features.

\subsection{The Machine Learning Classifier}
\begin{figure}[h!tbp]
\centering
\includegraphics[width=0.55\textwidth]{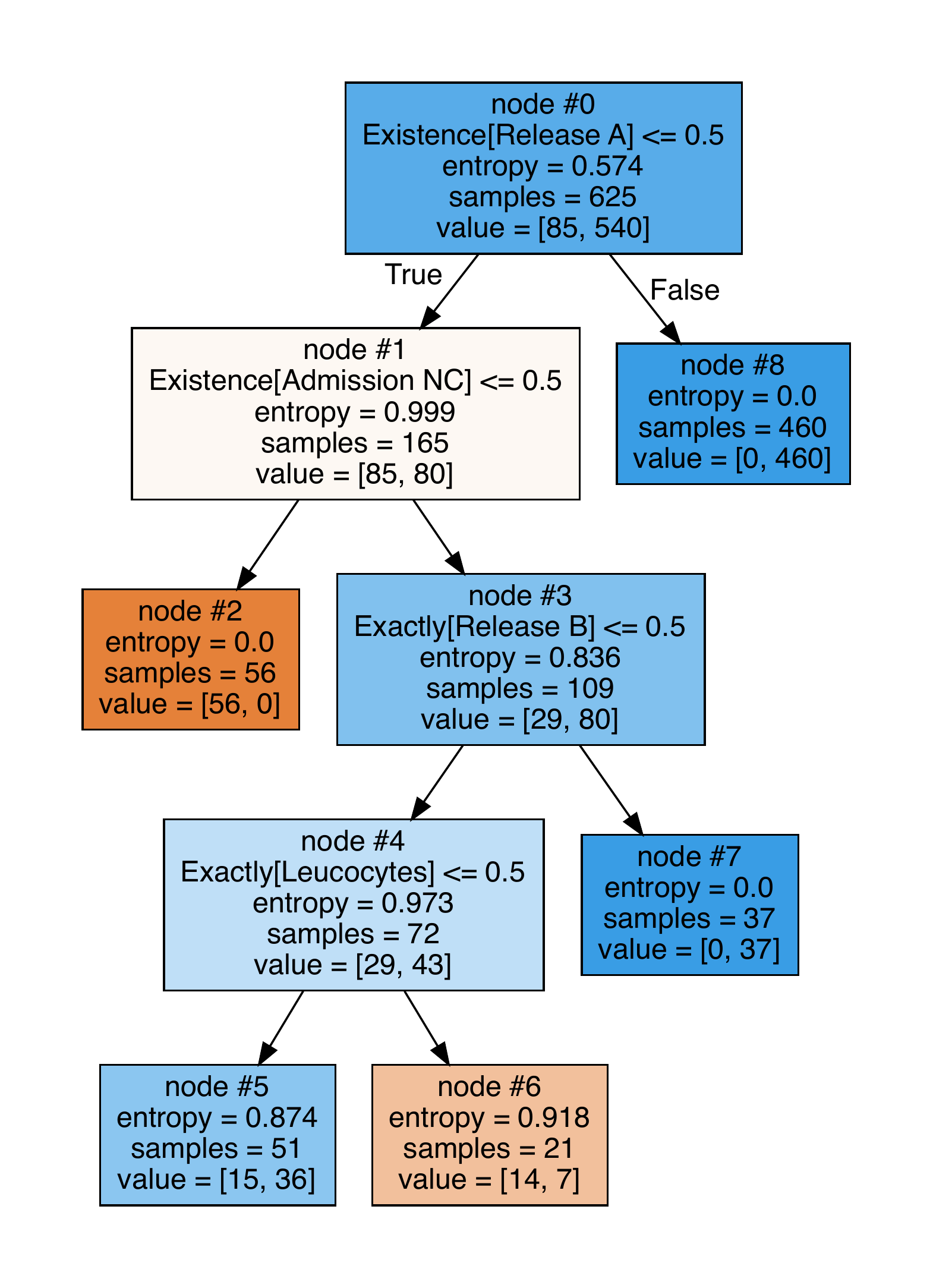}
\caption{Decision Tree for the $sepsis\_case\_2$ dataset. The blue and red leaves correspond to positive and negative paths, respectively. Left arrows correspond to conditions evaluated as true, right arrows correspond to conditions evaluated as false.}
\label{fig:dt_sepsis_2}
\end{figure}

The trained DT is shown in Figure~\ref{fig:dt_sepsis_2}.
The DT is relatively small with only 5 paths and a depth of 5. In spite of this, the DT does not underfit, but discriminates well between positive and negative samples as shown in Table~\ref{tab:DT_fscore} of Section 5.

By looking at the DT, there are 3 paths leading to a positive outcome. The most likely one $node_{\#0} \rightarrow node_{\#8}$ has maximal purity (1) and 85.2\% of the positive samples follow this path (460 out of 540). Also path $node_{\#0} \rightarrow node_{\#1} \rightarrow node_{\#3} \rightarrow node_{\#7}$ has maximal purity, although, in this case, only 6.9\% of the positive samples follow this path (37 out of 540). Finally, only 6.7\% of the positive samples follow the least likely path $node_{\#0} \rightarrow node_{\#1} \rightarrow node_{\#3} \rightarrow node_{\#4} \rightarrow node_{\#5}$, and this path has also a high entropy (0.874).

\subsection{Recommendation Generation}
The recommendations are generated using the recommendation score defined in Eq.~(\ref{eq:rho}). {The hyperparameters $\lambda_1, \lambda_2, \lambda_3$ that weight fitness, purity and positive sample probability in the definition of the recommendation score are found through grid search and their optimized values are, in our example, 0.4, 0.4 and 0.2. This means that fitness and purity have higher importance with respect to the positive sample probability.}

We now show some examples of recommendations generated starting from a given prefix and the above DT. For prefix $\sigma_{15} = \langle$\act{ER Sepsis Triage}, \act{ER Registration}, \act{ER Triage}, \act{CRP}, \act{LacticAcid}, \act{Leucocytes}, \act{IV Antibiotics}, \act{IV Liquid}, \act{Admission NC}, \act{CRP}, \act{Leucocytes}, \act{Admission NC}, \act{CRP}, \act{Leucocytes}, \act{Release B}$\rangle$, the positive path in the DT matching the prefix with the highest recommendation score is $node_{\#0} \rightarrow node_{\#1} \rightarrow node_{\#3} \rightarrow node_{\#7}$. {This path has a higher recommendation score with respect to the other positive paths as it has fitness and purity values equal to 1. The high value of the fitness function is due to the full compliance of the feature values in the encoding of $\sigma_{15}$ and the ones in the path. In particular, constraint \constr{existence} (\act{Release A}) is violated in the path and possibly violated in $\sigma_{15}$; \constr{existence} (\act{Admission NC}) is satisfied in both the path and in $\sigma_{15}$; \constr{exactly} (\act{Release B}) is satisfied in the path and possibly satisfied in $\sigma_{15}$. These high values counterbalance the low probability of the positive samples (1.3\%) in the computation of the recommendation score. 

The path having the highest positive sample probability ($node_{\#0} \rightarrow node_{\#8}$) has a lower fitness value (0.5) that leads to a lower recommendation score. The third path has a lower fitness value (0.875), lower purity and lower positive sample probability. Since constraint \constr{existence} (\act{Admission NC}) is already satisfied in prefix $\sigma_{15}$, the generated recommendations are} $\{\langle \text{\constr{existence} (\act{Release A})}$, $\text{It should not be SATISFIED}\rangle$, $ \langle \text{\constr{exactly} (\act{Release B})}$, $\text{It should not be VIOLATED}\rangle\}$. The first recommendation has a higher priority with respect to the second one. This recommendation suggests that {in order to avoid the intensive care, in case of hospitalization, activity \act{Release A} should not be performed. The second recommendation states, instead, that activity \act{Release B} has to be performed exactly once and, since this activity has already been performed in $\sigma_{15}$, it should not be performed again.} 

As a second example, we consider prefix $\sigma_{5} = \langle$\act{IV Liquid}, \act{ER Registration}, \act{ER Triage}, \act{ER Sepsis Triage}, \act{IV Antibiotics}$\rangle$. {In this case, the path with the highest recommendation score in the DT is $node_{\#0} \rightarrow node_{\#8}$. This path has high purity and positive sample probability (1 and 85.2\%, respectively) that counterbalance a modest fitness value (0.5). This fitness value is due to the low similarity of prefix $\sigma_5$ with the path (constraint \constr{existence} (\act{Release A}) is satisfied in the path but possibly violated in $\sigma_{5}$). The other positive paths have, instead, a higher fitness (0.67 for $node_{\#0} \rightarrow node_{\#1} \rightarrow node_{\#3} \rightarrow node_{\#7}$, and 0.875 for $node_{\#0} \rightarrow node_{\#1} \rightarrow node_{\#3} \rightarrow node_{\#4} \rightarrow node_{\#5}$) that, however, is not sufficient to counterbalance the low positive sample probability in the recommendation score. Therefore, in this case, only one recommendation is provided ($\{\langle \text{\constr{existence} (\act{Release A})}$, $\text{It should be SATISFIED}\rangle\}$), indicating that for a Sepsis case with a clinical history similar to $\sigma_{5}$, in order to avoid intensive care, activity \act{Release A} should occur at least once.

\section{Evaluation}
\label{sec:evaluation}

{To assess the validity of our proposal, we need to answer the following research questions:}
\begin{description}
\item[RQ1.] Are the recommendations extracted from a classifier trained on \declare constraints effective for achieving a desired outcome {in a business process execution?}
\item[RQ2.] Are there statistical differences in using different families of \declare constraints for extracting effective recommendations for a business process?
\end{description}

{To answer these research questions, on the one hand, we} check that the adoption of the recommendation set $\mathcal{R}$ brings to a positive outcome for a given trace of a business process. On the other hand, we {also}  show that if the recommendations are not followed, the process is going to achieve a negative outcome. 

{In particular, w}e developed the following experiment protocol on a pool of datasets to test the proposed Outcome-Oriented Prescriptive Process Monitoring system.
{For} each {event} log $\Log$ in the pool of datasets:
	\begin{enumerate}
	\item \textbf{[Preprocessing]} divide $\Log$ into $\Log_{train}$ and $\Log_{test}^*$;
	\item \textbf{[Preprocessing]} for each constraint family in {$\{\mathcal{E},  \mathcal{\widehat{C}},  \mathcal{\widehat{PR}},  \mathcal{\widehat{NR}},  \mathcal{A}\}$, where $\mathcal{\widehat{C}}=\mathcal{E} \cup \mathcal{C}$, $\mathcal{\widehat{PR}}=\mathcal{E} \cup \mathcal{PR}$ and  $\mathcal{\widehat{NR}}=\mathcal{E} \cup \mathcal{NR}$}:
		\begin{enumerate}
		\item \textbf{[Preprocessing]} use the \declare encoder to encode $\Log_{train}$ and $\Log_{test}^*$ according to the chosen constraint family (Section~\ref{sec:encoding});
		\item \textbf{[ML Classifier]} use the encoded log $\Log_{train}$ to train a DT $f_\theta$ (Section~\ref{sec:train});
		\item \textbf{[Recommendation Generation]} for each prefix length $k$ in $\Log_{test}^*$:
			\begin{enumerate}
			\item \textbf{[Recommendation Generation]} generate the recommendations $\mathcal{R}$ from $f_\theta$ for each encoded $\sigma_k \in \Log_{test}^*$ (Section~\ref{sec:rec_gen});
			\item \textbf{[Evaluation]} evaluate the quality of $\mathcal{R}$ on the whole trace $\sigma$ by considering $y(\sigma)$ (Section~\ref{sec:metrics}).
			\end{enumerate}
		\end{enumerate}
	\end{enumerate}


\subsection{Datasets}
\label{sec:datasets}
As a pool of datasets, we adopt the one used in~\cite{TeinemaaDRM19} used as a benchmark for Outcome-oriented Predictive Process Monitoring. Such well-known and standard datasets allow us to a have a robust and significant evaluation of our Prescriptive Process Monitoring system. Following~\cite{TeinemaaDRM19}, we used eight real-life event logs publicly available in the 4TU Centre for Research Data\footnote{\url{https://data.4tu.nl/repository/collection:event_logs_real}} and discarded the private \emph{Insurance} dataset (since it is not publicly available). In most of the datasets, several labeling functions $y$ have been applied, i.e., different desired outcomes in each dataset are specified.
These labelings on the eight initial event logs lead to 22 different prescriptive tasks and datasets. We now provide more details on the original logs, the used labeling functions and the resulting Prescriptive Process Monitoring tasks.

{\bfseries BPIC 2011.} This event log has been originally published in relation to the Business Process Intelligence Challenge (BPIC) that took place in 2011. This event log {refers to} cases from the Gynaecology department of a Dutch Academic Hospital. Each case 
records procedures and treatments (stored as activities) applied to a given patient. There are four different labeling functions based on {four} LTL formulas~\cite{pnueli1977temporal}, that is, the class label for a case $\sigma$ is defined according to the satisfaction of {the} LTL formula $\varphi$ in each trace $\sigma$:
\begin{equation*}
 y(\sigma) =
 \begin{cases}
   1  & \text{if } \varphi \text{ violated in } \sigma \\
   0  & \text{otherwise}
 \end{cases}
\end{equation*}
The four LTL rules used are the following:
\begin{itemize}
\begin{footnotesize}
  \item \emph{bpic2011\_1}: $\varphi =  \lfuture(\act{tumor~marker~CA-19.9}) \vee  \lfuture(\act{{ca-125}~using~meia})$;
  \item \emph{bpic2011\_2}: $\varphi =  \lglobally(\act{CEA-tumor~marker~using~meia} \rightarrow$ $\lfuture(\act{squamous~cell~carcinoma~using~meia}))$;
  \item \emph{bpic2011\_3}: $\varphi =  (\neg \act{histological~examination-biopsies~nno})$ $\luntil (\act{squamous~cell~carcinoma~using~meia})$;
  \item \emph{bpic2011\_4}: $\varphi =  \lfuture(\act{histological~examination-big~resectiep})$.
 \end{footnotesize}
\end{itemize}

For example, the labeling for \emph{bpic2011\_1} expresses the fact that at least one of the activities \act{tumor~marker} \act{CA-19.9} or \act{{ca-125}~using~meia} must happen eventually during a case. It is trivial to see that when one of these events occur, the class label  $y(\sigma)$ becomes known. Therefore, the evaluation step will be biased due to this phenomenon. To solve this issue, all the cases have been cut exactly before the occurrence of one of these events. The same cut is performed exactly before the occurrence of \act{histological~examination-biopsies~nno} in \emph{bpic2011\_3} and before \act{histological~examination-big~resectiep} in \emph{bpic2011\_4}. Regarding \emph{bpic2011\_2}, no cut is necessary as it is never possible to infer the class label before the end of the case. Indeed, the class label is true if and only if every occurrence of \act{CEA-tumor~marker~using~meia} is eventually followed by \act{squamous~cell~carcinoma~using~meia} and this constraint is never permanently satisfied or violated before the end of the case.

{\bfseries BPIC 2012.} This {event log} {refers to} the execution history of a loan application process in a Dutch Financial Institution. Each case stores the events related to a particular loan application. The available labelings are based on the final outcome of a loan application, i.e., on whether the application is accepted, rejected, or canceled. This is a multi-class classification problem, but, as in~\cite{TeinemaaDRM19}, the labelings are considered as three separate binary classification tasks. In the experiments, these tasks are referred to as \emph{bpic2012\_accepted}, \emph{bpic2012\_cancelled}, and \emph{bpic2012\_refused}.

{\bfseries BPIC 2015.} This {event log} {refers to} 
the application process of building permits of 5 Dutch Municipalities. Each log comes from a single Municipality and is taken as a single dataset with its own labeling function. This is defined similarly to BPIC 2011, that is, according to the satisfaction/violation of an LTL formula $\varphi$. Each dataset is denoted as \emph{bpic2015\_i}, where $i = 1 \ldots 5$ indicates the number of the Municipality. The adopted labeling function is:
\begin{itemize}
\item \emph{bpic2015\_i}: $\varphi = \lglobally(\act{send~confirmation~receipt} \rightarrow \lfuture(\act{retrieve~missing~data})) $.
\end{itemize}
Similarly to \emph{bpic2011\_2}, no trace cutting has been performed as the satisfaction/violation of $\varphi$ can be evaluated only at the completion of the case.

{\bfseries BPIC 2017.} This {event log} originates from the same Financial Institution as $bpic2012$, but with an improvement of the data collection process, resulting in a richer and cleaner dataset. As for $bpic2012$, the event cases record execution traces of a loan application process {and} 
three separate labelings based on the outcome of the application are applied, i.e., \emph{bpic2017\_accepted}, \emph{bpic2017\_cancelled}, and \emph{bpic2017\_refused}.

{\bfseries Hospital billing.} This dataset contains cases regarding a billing procedure for medical services. The cases come from an ERP system of a Hospital and the labelings for this {log} 
are:
\begin{itemize}
\item \emph{hospital\_1}: the billing procedure is not eventually closed;
\item \emph{hospital\_2}: the billing procedure is reopened.
\end{itemize}

{\bfseries Production.} This {event log} contains cases of a manufacturing process. Each case stores information about the activities, workers and/or machines involved in the production process of an item. The labeling is based on whether, in a case, there are rejected work orders, or not.

{\bfseries Sepsis cases.} This dataset records hospitalizations of patients with symptoms of the life-threatening Sepsis condition in a Dutch Hospital. Each case stores events from the patient's registration in the Emergency Room (\act{ER registration}) to the discharge from the Hospital. Laboratory tests together with their results are also recorded as events. The reasons of the discharge are available in an anonymized format. Three different labelings for this log are available:
\begin{itemize}
\item \emph{sepsis\_1}: the patient returns to the Emergency Room within 28 days from the discharge;
\item \emph{sepsis\_2}: the patient is (eventually) admitted to intensive care;
\item \emph{sepsis\_3}: the patient is discharged from the Hospital on the basis of a reason different from \emph{Release A} (i.e., the most common release type).
\end{itemize}

{\bfseries Traffic fines.} This event log comes from the ERP of an Italian local Police Force. The events in the log refer to the notifications sent about a fine and the (partial) repayments. Additional case/event attributes include, for instance, the reason, the total amount, and the amount of repayments for each fine. The available labeling is based on whether the fine is repaid in full, or is sent for credit collection.

The adopted 22 datasets exhibit different characteristics shown in Table~ \ref{tab:dataset_stats}. The \emph{production} log is the smallest one with 220 cases, while the \emph{traffic} log is the largest one with 129\,615 cases. The datasets with the highest case lengths are the \emph{bpic2011} datasets where the longest case has 1814 events. On the other hand, the \emph{traffic} log contains the shortest cases (their length varies from 2 to 20 events). The class labels are the most imbalanced in the \emph{hospital\_billing\_2} dataset, where only 5\% of cases are labeled as \emph{positive} (class label = 1). Conversely, in the \emph{bpic2012\_accepted}, \emph{bpic2017\_cancelled} and \emph{traffic} datasets, the classes are balanced. {Concerning} the event classes, \emph{traffic\_fines\_1} has the lowest number of distinct activity names (10). On the other hand, the logs with the highest number of event classes are the \emph{bpic2015} logs containing a maximum of 396 event classes.
\begin{table}[h!t]
\caption{Statistics of the datasets used in the experiments.}
\label{tab:dataset_stats}
\centering
\scalebox{0.8}{
\begin{tabular}{lcccccc}
\toprule
\textbf{Dataset}     & \textbf{\makecell{Cases\\Count}} & \textbf{\makecell{Min\\Length}} & \textbf{\makecell{Median\\Length}} & \textbf{\makecell{Max\\Length}} & \textbf{\makecell{Positive Cases\\Ratio}} & 
\textbf{\makecell{Event Classes\\Count}} \\
\midrule
$bpic2011\_1$         & 1140                & 1                   & 25.0                   & 1814                & 0.40 & 193                     \\
$bpic2011\_2$         & 1140                & 1                   & 54.5                   & 1814                & 0.78        & 251              \\
$bpic2011\_3$         & 1121                & 1                   & 21.0                   & 1368                & 0.23           &190           \\
$bpic2011\_4$         & 1140                & 1                   & 44.0                   & 1432                & 0.28        &231              \\
$bpic2012\_cancelled$  & 4685                & 15                  & 35.0                   & 175                 & 0.35         &36             \\
$bpic2012\_accepted$   & 4685                & 15                  & 35.0                   & 175                 & 0.48        &36              \\
$bpic2012\_rejected$   & 4685                & 15                  & 35.0                   & 175                 & 0.17        &36              \\
$bpic2015\_1$      & 696                 & 2                   & 42.0                   & 101                 & 0.23       &380               \\
$bpic2015\_2$      & 753                 & 1                   & 55.0                   & 132                 & 0.19      & 396                \\
$bpic2015\_3$      & 1328                & 3                   & 42.0                   & 124                 & 0.20      &380                \\
$bpic2015\_4$      & 577                 & 1                   & 42.0                   & 82                  & 0.16     &319                 \\
$bpic2015\_5$      & 1051                & 5                   & 50.0                   & 134                 & 0.31        & 376              \\
$bpic2017\_accepted$   & 31\,413               & 10                  & 35.0                   & 180                 & 0.41      &26                \\
$bpic2017\_cancelled$  & 31\,413               & 10                  & 35.0                   & 180                 & 0.47      &26                \\
$bpic2017\_rejected$    & 31\,413               & 10                  & 35.0                   & 180                 & 0.12       &26               \\
$hospital\_billing\_1$ & 77\,525               & 2                   & 6.0                    & 217                 & 0.10         &18             \\
$hospital\_billing\_2$ & 77\,525               & 2                   & 6.0                    & 217                 & 0.05         &17             \\
$production$           & 220                 & 1                   & 9.0                    & 78                  & 0.53          &26            \\
$sepsis\_cases\_1$     & 782                 & 5                   & 14.0                   & 185                 & 0.14        &24              \\
$sepsis\_cases\_2$     & 782                 & 4                   & 13.0                   & 60                  & 0.14        &24              \\
$sepsis\_cases\_3$     & 782                 & 4                   & 13.0                   & 185                 & 0.86           &24           \\
$traffic\_fines$    & 129\,615              & 2                   & 4.0                    & 20                  & 0.46              &10       \\
\bottomrule
\end{tabular}
}
\end{table}

\revision{These datasets are standard benchmarks that do not require particular cleaning operations as preprocessing. The preprocessing is limited to the \declare encoding of the traces and to the removal of too long traces that could bias the evaluation (see Section~\ref{sec:setup}).}

\subsection{Offline Evaluation of a Prescriptive Process Monitoring System}
\label{sec:metrics}
{One of the main challenges when evaluating Prescriptive Process Monitoring systems is dealing with the lack of adoption of those systems by real users~\cite{Dumas21}. One of the possibilities when testing these systems is, therefore, to resort to an offline evaluation based on the ``what-if'' simulation~\cite{Dumas21}, in order to evaluate the effectiveness of the set of  recommendations $\mathcal{R}$ for a prefix $\sigma_k$.
The idea is evaluating the consequences of (not) following the recommendations $\mathcal{R}$ at step $k$ on the whole trace $\sigma$. We hence evaluate the effectiveness of $\mathcal{R}$ for a prefix $\sigma_k$ by checking whether the recommendations in $\mathcal{R}$ have been followed in $\sigma$, and by comparing the outcome of $\sigma$ and its actual label $y(\sigma)$. We expect that if the recommendations are followed, the outcome will be positive. If they are not followed, the outcome will be negative.}
Let $p^*$ be the path of the DT from which {the set} $\mathcal{R}$ has been computed. A high similarity between $\sigma$ and $p^*$ means {that the recommendations have been followed by the execution $\sigma$ and hence that we expect a positive outcome. The prediction related to trace $\sigma$ will hence be classified as a \emph{true positive (TP)} if $y(\sigma) = 1$ or as a \emph{false positive (FP)} if $y(\sigma) = 0$. Symmetrically, if there is no similarity between $\sigma$ and $p^*$, this means that the recommendations have not been followed by $\sigma$. We hence expect a negative outcome. The prediction related to trace $\sigma$ will hence be classified as a \emph{true negative (TN)} if $y(\sigma) = 0$ and as a \emph{false negative (FN)} if $y(\sigma) = 1$. 

The similarity between $\sigma$ and $p^*$ is computed by leveraging $\mathcal{F}(\sigma, p^*)$ (Eq.~(\ref{eq:fitness})). Differently from the general formula, however, in this case the compliance function $\mathcal{C}$ is applied to the whole trace and, therefore, does not need to take into account temporary violations/satisfactions of \declare constraints in $p^*$.
Specifically, a fitness threshold $th_{fit}$ is used to evaluate the similarity between the whole trace $\sigma$ and the path $p*$, that is, if  $\mathcal{F}(\sigma, p^*)$ is higher than or equal to $th_{fit}$, this means that the recommendations in $\mathcal{R}$ are followed. A similarity lower than $th_{fit}$ means that the trace did not follow the recommendations. Adopting a fitness threshold is necessary as a similarity of exactly 1 between $\sigma$ and $p^*$ could be too restrictive and lead to a high number of false negatives. This is totally in line with a realistic situation in which some of the recommendations are not followed by a process manager as they are not strictly necessary for the positive outcome of the process.
%
In our experiments, the optimal fitness thresholds have been selected via grid search. Table~\ref{tab:confusion_matrix} summarizes the confusion matrix entries.}

\begin{table}
\begin{center}
\begin{tabular}{l|c|c|}
\toprule
                & $\mathcal{F}(\sigma, p^*) \geq th_{fit}$ & $\mathcal{F}(\sigma, p^*) < th_{fit}$ \\
                \midrule
$y(\sigma) = 1$ & $TP$                                  & $FN$                                  \\\midrule
$y(\sigma) = 0$ & $FP$                                  & $TN$                   \\
\bottomrule              
\end{tabular}
\caption{{Confusion matrix for the evaluation of the recommendations.}}
\label{tab:confusion_matrix}
\end{center}
\end{table}


{To assess the accuracy of our approach, we compute precision, recall and F-score as follows:

\begin{equation*}
prec = \frac{TP}{TP + FP}; \qquad\quad  rec = \frac{TP}{TP + FN}; \qquad\quad  F-score = \frac{2*prec*rec}{prec + rec}.
\end{equation*}

We use F-score rather than accuracy as many of the datasets used in the evaluation are imbalanced towards the negative class and the accuracy could be biased by the true negatives leading to non-reliable results.}

\subsection{Experimental Setup}
\label{sec:setup}
In this section, we provide some details about the experimental setup. All the experiments were carried out using Python 3.6, the \textsc{Declare4Py} library~\cite{DonadelloRMS22} (for the \declare encoding of the traces) and the scikit-learn library 0.24~\cite{scikit-learn} (for building and querying the classifiers).\secRevision{ We also provide a link\footnote{\url{https://github.com/ivanDonadello/LTL-prescriptive-process-monitoring}} to an online repository containing the source code of the experiments along with the link to the datasets, the trained decision trees and the optimized values of the hyperparameters.}
\paragraph{\textbf{Preprocessing}}
Mimicking real-life situations in which the prediction model is trained on historical data and the recommendation is carried out on ongoing cases, the event logs have been first chronologically ordered and then split in training and test set. Specifically, the cases in the event logs have been ordered according to the start time and the first 80\% -- i.e., all cases that started before a given date -- has been used for the construction of the training and the validation log, while the remaining 20\% has been used to create the test event log $\Log_{test}^*$. Since the last cases of the training and validation log could still not be completed when the test period starts, we removed from these cases in the training and validation log the events overlapping with the test period, as in~\cite{TeinemaaDRM19}. The training and the validation event logs are instead split so that the first 70\% of the whole event log ($\Log_{train}$) is used for training the prediction model, while about 10\% of the event log ($\Log_{val}$) is used for the optimization of the hyperparameters. \revision{Another preprocessing operation is the removal of too long traces that could bias the evaluation. This operation is better explained in the following.}

\paragraph{\textbf{ML Classifier Training}}
The training of the DT has been performed with a grid search to tune the hyperparameters with 5-fold cross-validation {on $\Log_{train}$}. The range of values used for the hyperparameters are: i) the Gini index or the entropy criterion for the computation of the impurity; ii) $[4, 6, 8, 10, \infty]$ for the maximum depth of the DT; iii) the use of class weights or not during the training to avoid poor performance due to the imbalance of the datasets (see, for example, $hospital\_billing\_2$ in Table~\ref{tab:dataset_stats}); iv) $[0.1, 0.2, 0.3, 2]$ for the minimum number of samples required to split an internal node (float values indicate a percentage of the training data); v) $[1, 10, 16]$ for the minimum number of samples required to consider a node a leaf node; vi) the number of the most informative features to use in the feature selection phase, i.e., 50\%, 30\% and the square root of the total number of initial features (after ranking them by using the mutual information score).
\paragraph{\textbf{Recommendation Generation}}
Using the trained DT, the $\lambda$ parameters in Eq.~\eqref{eq:rho} have been optimized through grid search on the prefix log $\Log^*_{val}$ extracted from the validation log $\Log_{val}$. The set $\mathcal{P}^+$ in the same equation has been filtered to contain only paths with at least 3 training samples.
\paragraph{\textbf{Evaluation}}
The values $k$ for the prefix lengths range from 1 to a maximum that changes according to the dataset. We adopted the same criteria used in~\cite{TeinemaaDRM19} for the maximum value: 9 for the $traffic\_fines$ dataset, the minimum between 20 and the 90\textsuperscript{th} percentile of the case lengths for the $bpic2017$ datasets, the minimum between 40 and the 90\textsuperscript{th} percentile of the case lengths for the other datasets. This choice is due to the low number of long cases (after the 90\textsuperscript{th} percentile) in the prefix test logs that could produce results with no statistical significance for high values of $k$. The optimal fitness threshold $th_{fit}$ has been found by {applying} grid search on $\Log_{val}^*$ and using values 0.55, 0.65, 0.75, 0.85. These values have been chosen considering that values outside this range could bias the system towards a low precision or a low recall.

\subsection{Results}
\label{sec:results}
{Since the proposed approach leverages a DT trained on $\Log_{train}$ to provide recommendations, we first inspect the performance of the DT in the classification of the outcome of (complete) traces in $\Log_{val}$ and $\Log_{train}$ as positive or negative.}
Table \ref{tab:DT_fscore} shows the average F-score of the DT on $\Log_{val}$ and $\Log_{train}$.
\begin{table*}[htbp]
\caption{DTs F-score (validation/train).}
\label{tab:DT_fscore}
\begin{tabular}{lccccc}
\toprule
\textbf{Dataset}     & \textbf{$\mathcal{E}$} & \textbf{$\widehat{\mathcal{C}}$} & \textbf{$\widehat{\mathcal{PR}}$}  & \textbf{$\widehat{\mathcal{NR}}$} & \textbf{$\mathcal{A}$} \\

\midrule
$bpic2011\_1$ & 95.21/95.24 & 95.45/95.52 & 98.1/98.1 & 95.69/95.71 & 98.1/98.1 \\ 
$bpic2011\_2$ & 98.74/98.74 & 98.86/99.08 & 99.06/99.06 & 96.95/97.25 & 99.0/99.06 \\ 
$bpic2011\_3$ & 90.32/98.29 & 91.69/91.83 & 97.98/98.0 & 90.99/96.53 & 97.98/98.0 \\ 
$bpic2011\_4$ & 84.3/94.01 & 83.1/83.89 & 95.42/95.98 & 87.03/97.06 & 94.88/94.92 \\ 
$bpic2012\_accepted$ & 92.09/93.42 & 92.93/93.44 & 92.67/93.75 & 88.37/91.13 & 88.37/91.16 \\ 
$bpic2012\_cancelled$ & 92.44/90.74 & 92.4/90.77 & 91.86/91.61 & 92.4/90.77 & 92.43/91.59 \\ 
$bpic2012\_rejected$ & 86.65/90.22 & 86.65/90.22 & 86.65/90.22 & 86.65/90.22 & 86.65/90.06 \\ 
$bpic2015\_1$ & 96.5/98.07 & 96.89/97.74 & 96.5/97.97 & 96.89/97.97 & 96.5/98.38 \\ 
$bpic2015\_2$ & 96.04/96.55 & 95.63/97.51 & 97.84/97.89 & 96.98/98.76 & 96.08/97.89 \\ 
$bpic2015\_3$ & 98.07/98.49 & 98.07/98.69 & 96.8/98.49 & 98.07/99.22 & 98.07/98.69 \\ 
$bpic2015\_4$ & 96.64/98.53 & 97.93/98.69 & 97.28/98.69 & 97.93/98.69 & 97.93/98.69 \\ 
$bpic2015\_5$ & 99.06/99.22 & 98.86/99.31 & 99.07/99.4 & 99.06/99.07 & 98.68/99.82 \\ 
$bpic2017\_accepted$ & 93.36/93.88 & 94.61/95.88 & 96.25/97.68 & 96.24/97.67 & 96.27/97.62 \\ 
$bpic2017\_cancelled$ & 95.32/96.12 & 95.79/95.69 & 97.3/97.49 & 96.96/97.67 & 97.39/97.51 \\ 
$bpic2017\_rejected$ & 95.83/96.06 & 95.82/94.23 & 95.92/96.37 & 95.83/96.06 & 95.82/94.23 \\ 
$hospital\_billing\_1$ & 91.63/91.82 & 91.65/91.82 & 91.04/91.2 & 91.04/91.2 & 91.04/91.2 \\ 
$hospital\_billing\_2$ & 57.13/55.76 & 54.4/53.4 & 57.3/57.88 & 29.19/24.4 & 54.42/53.44 \\ 
$production$ & 83.11/89.07 & 83.32/94.89 & 84.94/87.07 & 84.92/84.64 & 82.23/91.52 \\ 
$sepsis\_cases\_1$ & 28.61/31.32 & 28.24/28.92 & 30.58/39.56 & 33.55/34.37 & 30.58/43.05 \\ 
$sepsis\_cases\_2$ & 79.04/79.41 & 79.04/79.41 & 78.5/80.26 & 79.04/80.26 & 90.28/89.91 \\ 
$sepsis\_cases\_3$ & 96.42/96.42 & 96.42/96.42 & 96.42/96.5 & 96.42/96.42 & 96.42/96.5 \\ 
$traffic\_fines$ & 92.32/94.37 & 93.42/94.37 & 92.92/94.96 & 95.4/96.38 & 92.92/94.96 \\    
\bottomrule
\end{tabular}
\end{table*}
We notice that the majority of the classifiers have good performance on both the validation and train folds with an absence (or a low degree) of overfitting. However, the $hospital\_billing\_2$ and the $sepsis\_cases\_1$ datasets present low performance on both $\Log_{val}$ and $\Log_{train}$. This underfitting is due to the {insufficient information} carried by the adopted encoding. We checked this {aspect} by inspecting the traces of these datasets with the Disco\footnote{https://fluxicon.com/disco/} tool. We noticed that both positive and negative labeled traces present a very similar control flow. Therefore, any ML classifier taking as input traces encoded with our \declare-based encoding do{es} not have sufficient information for discriminating between positive and negative samples. As future work, we aim at enriching our encoding with information regarding the data payloads attached to events and their execution times to overcome this issue.
In general, no significant difference between families of \declare constraints used for trace encoding is found in these results. Only the $sepsis\_cases\_2$ dataset benefits of a higher number of features provided in the $\mathcal{A}$ family.

We now discuss the results {related to the returned recommendations.}
Figures \ref{fig:rec_res_1} and \ref{fig:rec_res_2} show the trend of the F-score {for different prefix lengths $k$ computed on each prefix  $\sigma_k$ in $\Log_{test}^*$.}
Both figures show the cumulative results, that is, the $TP$, $TN$, $FP$ and $FN$ at a given prefix length $k$ are summed {to the corresponding ones at prefix $j \le k$.}
This is done to avoid that the results {are} influenced by {the} small number of traces that {usually characterize long traces~\cite{TeinemaaDRM19}.}
{The average over all the prefixes is reported in Table~\ref{tab:avg_fscore}.}
\begin{figure}[h!t]
  \centering
  \includegraphics[width=\linewidth]{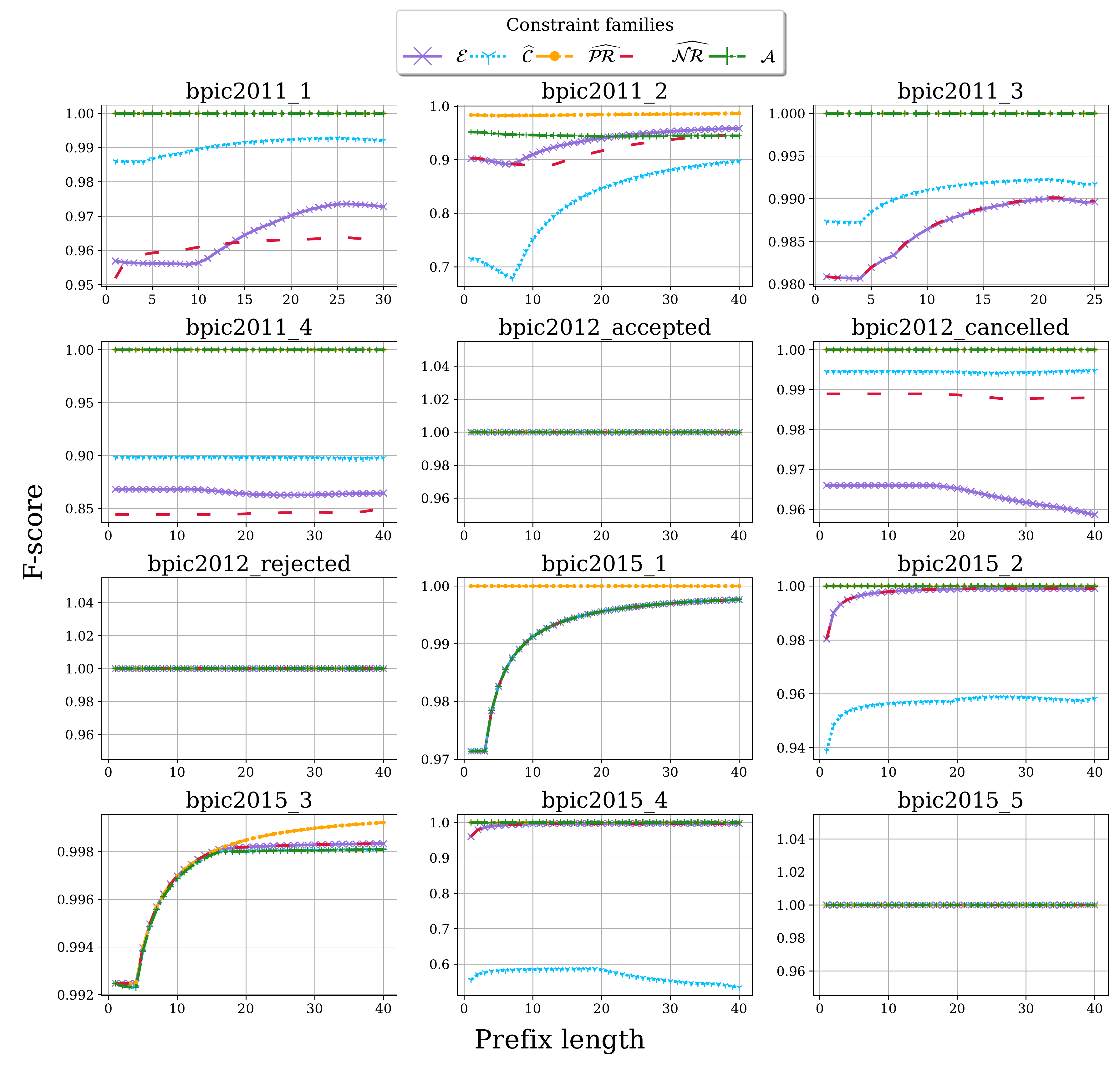}
  \caption{The cumulative F-score over all prefixes.}
  \label{fig:rec_res_1}
\end{figure}
\begin{figure}[h!t]
  \centering
  \includegraphics[width=\linewidth]{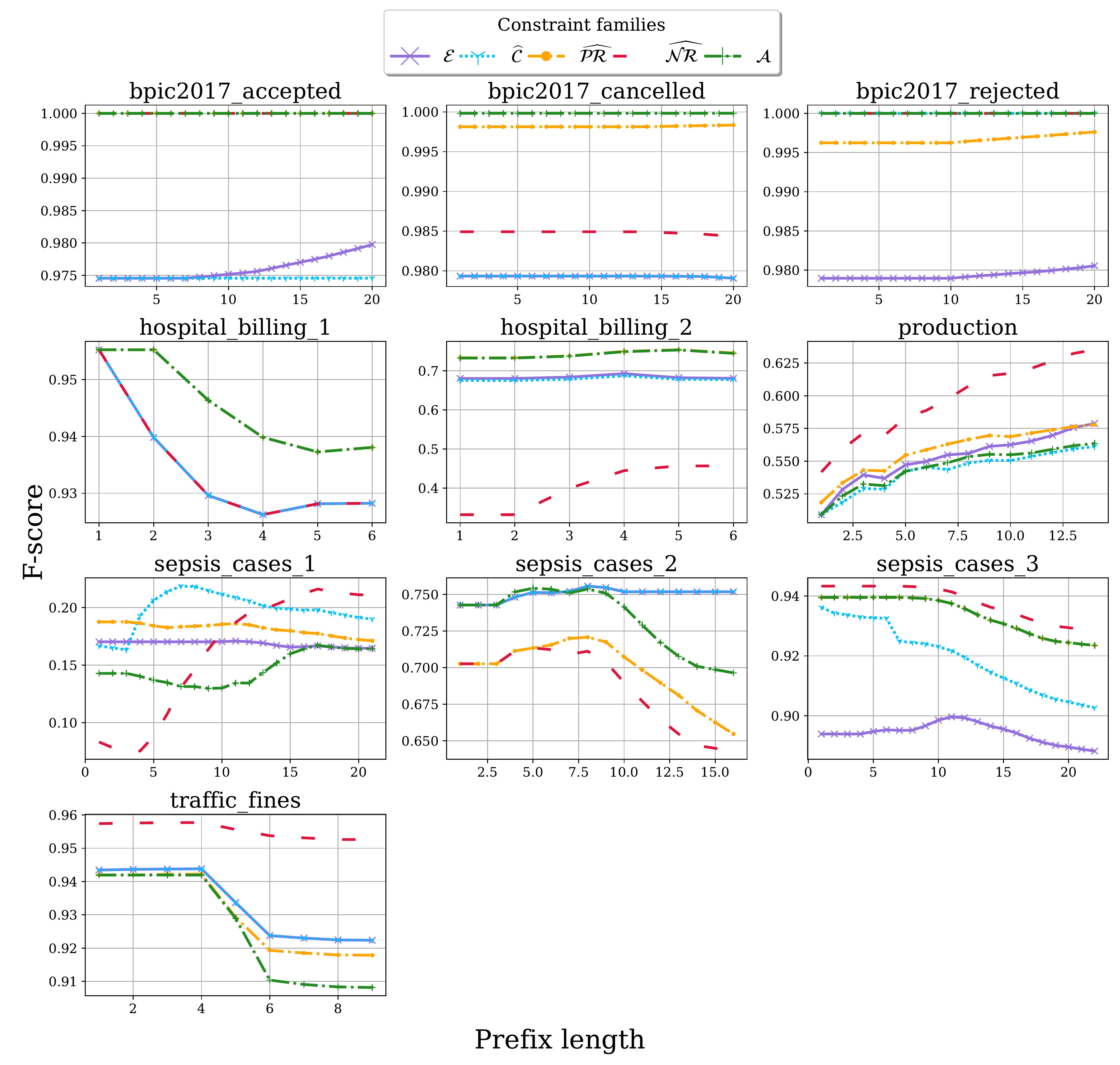}
  \caption{The cumulative F-score over all prefixes (continued).}
  \label{fig:rec_res_2}
\end{figure}
\begin{table*}
\caption{Average F-score over all prefixes for each family of \declare constraints. The best results are in bold.}
\label{tab:avg_fscore}
\begin{tabular}{lccccc}
\toprule
\textbf{Dataset}     & \textbf{$\mathcal{E}$} & \textbf{$\widehat{\mathcal{C}}$} & \textbf{$\widehat{\mathcal{PR}}$}  & \textbf{$\widehat{\mathcal{NR}}$} & \textbf{$\mathcal{A}$} \\
\midrule
$bpic2011\_1$         & 96.46     & 99.04  & \textbf{100.0}              & 96.15              & \textbf{100.0} \\
$bpic2011\_2$         & 93.32     & 81.98  & \textbf{98.42}              & 91.74              & 94.54 \\
$bpic2011\_3$         & 98.66     & 99.07  & \textbf{100.0}              & 98.67              & \textbf{100.0} \\
$bpic2011\_4$         & 86.53     & 89.81  & \textbf{100.0}              & 84.54              & \textbf{100.0} \\
$bpic2012\_accepted$  & \textbf{100.0}     & \textbf{100.0}  & \textbf{100.0}              & \textbf{100.0}              & \textbf{100.0} \\
$bpic2012\_cancelled$ & 96.38     & 99.44  & \textbf{100.0}              & 98.84              & \textbf{100.0} \\
$bpic2012\_rejected$  & \textbf{100.0}     & \textbf{100.0}  & \textbf{100.0}              & \textbf{100.0}              & \textbf{100.0} \\
$bpic2015\_1$         & 99.25     & 99.25  & \textbf{100.0}              & 99.25              & 99.25 \\
$bpic2015\_2$         & 99.77     & 95.65  & \textbf{100.0}              & 99.77              & \textbf{100.0} \\
$bpic2015\_3$         & 99.72     & 99.71  & \textbf{99.76}              & 99.72              & 99.71 \\
$bpic2015\_4$         & 99.51     & 56.79  & \textbf{100.0}              & 99.51              & \textbf{100.0} \\
$bpic2015\_5$         & \textbf{100.0}     & \textbf{100.0}  & \textbf{100.0}              & \textbf{100.0}              & \textbf{100.0} \\
$bpic2017\_accepted$  & 97.6      & 97.46  & \textbf{100.0}              & \textbf{100.0}              & \textbf{100.0} \\
$bpic2017\_cancelled$ & 97.93     & 97.93  & 99.82              & 98.48              & \textbf{99.98} \\
$bpic2017\_rejected$  & 97.94     & \textbf{100.0}  & 99.66              & \textbf{100.0}              & \textbf{100.0} \\
$hospital\_billing\_1$ & 93.46     & 93.46  & \textbf{94.53}              & 93.46              & \textbf{94.53} \\
$hospital\_billing\_2$ & 68.34     & 67.85  & \textbf{74.23}              & 40.33              & 74.17 \\
$production$         & 55.25     & 54.26  & 55.84              & \textbf{59.76}              & 54.55 \\
$sepsis\_cases\_1$     & 16.85     & \textbf{19.76}  & 18.16              & 16.17              & 14.66 \\
$sepsis\_cases\_2$     & \textbf{75.03}     & \textbf{75.03}  & 69.82              & 68.69              & 73.35 \\
$sepsis\_cases\_3$     & 89.43     & 91.94  & 93.37              & \textbf{93.8}               & 93.37 \\
$traffic\_fines$      & 93.33     & 93.33  & 93.01              & \textbf{95.53}              & 92.58 \\
\bottomrule
\end{tabular}
\end{table*}
Both the figures and the table show very good results in general, that is, the proposed Prescriptive Process Monitoring system returns recommendations that guarantee {a positive outcome in a trace. When the recommendations are not followed, instead, the corresponding traces have, in most of the cases, a negative outcome (\textbf{RQ1}). }


For some datasets, our system reaches an F-score of 100\% along all prefixes. This is due to the encoding with \declare patterns that creates a semantically rich feature space {that allows a crisp discrimination between regions containing only positive and regions containing only negative samples.} 
In addition, the resulting DTs have a depth of maximum 4 with a consequent 
lower number of temporal relations to {satisfy}. Therefore, the fitness and the overall performance of the system increase.

Although, in most of the cases, the high discriminating power of the temporal constraints on the control flow obtained with the \declare encoding guarantees accurate results, for some logs, the \declare encoding does not achieve such a good performance. The $production$ dataset, for instance, has a DT with depth 8 and contains several paths. In this situation, finding the best path is harder, the fitness score that can be achieved is lower and, as a consequence, the overall performance of the Prescriptive Process Monitoring system decreases. Moreover, our system poorly performs on the $sepsis\_cases\_1$ dataset due to the poor performance obtained by the DT (see Table~\ref{tab:DT_fscore}).

{
We stress the fact that other encodings, like the ones used in \cite{TeinemaaDRM19}, are based on a fine-grained vectorization of the log traces that minimizes the loss of information in each trace. For example, the well-known index encoding~\cite{LeontjevaCFDM15}} assigns at position $i$ of the feature vector the activity name of the event occurring at position $i$ in the trace. Therefore, the inference of high-level relations in the control flow of a trace, such as \constr{response} (\act{A}, \act{B}) or \constr{existence3} (\act{A}), is left to the ML system. In this case, the semantics inference is limited by the expressive capabilities of the ML system being used. The \declare encoding, on the other hand, is less fine grained as it abstracts the temporal order of the events with the constraint families. Since these relations are explicitly defined, the ML system only needs to infer the correlation between such the temporal patterns and the trace labeling. For this reason, using the \declare encoding, also simple ML models (like DTs) can easily capture those correlations thus improving their performance.


Since the use of \declare patterns as features in {Predictive and Prescriptive Process Monitoring}
is at its early adoption~\cite{dfmMaking2022},
we are interested in studying the effect of the different families of constraints on the quality of the recommendations. \secRevision{We, therefore, performed the Friedman's test~\cite{JamaludinRKBMM22,math10060915} on the columns of Table~\ref{tab:avg_fscore} to understand whether there is a statistically significant difference when using different families of constraints for generating the recommendations. We found a p-value of 0.0007 that rejects the null hypothesis and confirms the different impact of different families of constraints. However, we are also interested in which specific constraint family impacts the most.} Figure~\ref{fig:cd} is derived from Table~\ref{tab:avg_fscore} and shows the critical difference diagram of the \declare families by using the Nemenyi test with a significance level of 0.05 (as proposed in~\cite{Demsar06}).
\begin{figure}[h!t]
\centering
\includegraphics[width=\textwidth]{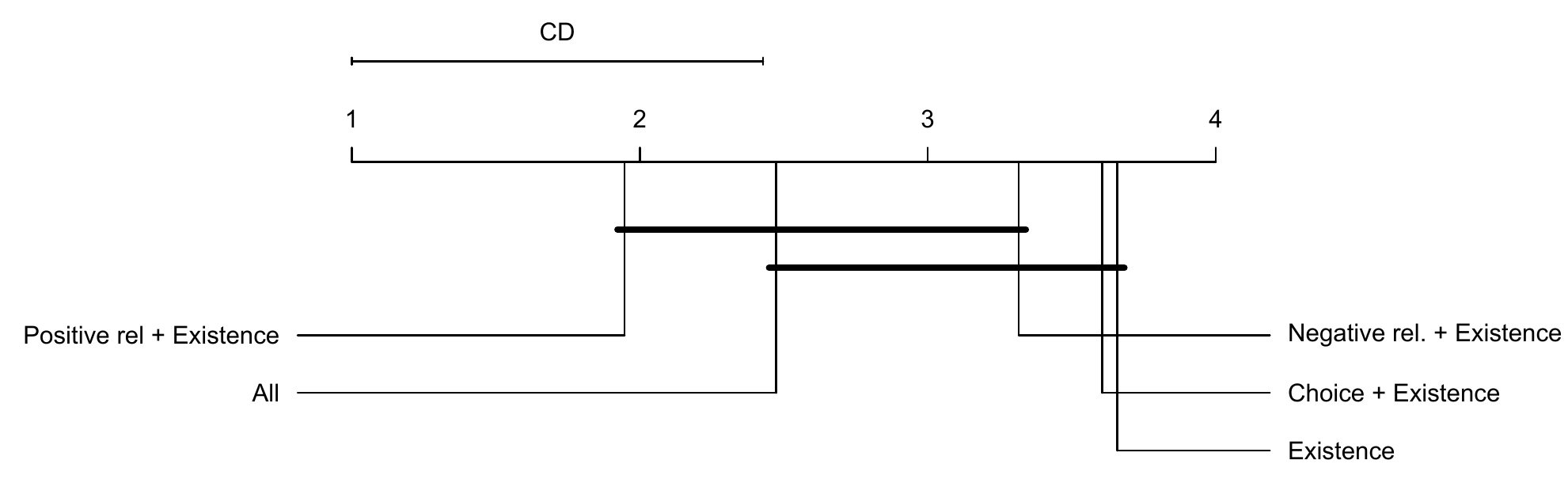}
\caption{Comparison of the \declare families with the Nemenyi test. The constraint families are compared in terms of the best F-score achieved in the datasets. The $\mathcal{\widehat{PR}}$ family achieves the best results in most of the datasets.}
\label{fig:cd}
\end{figure}
The diagram reports the average ranking of each family according to the F-score results in Table~\ref{tab:avg_fscore}. Groups of families that are not significantly different (with $p < 0.05$) are connected. We can observe that the $\mathcal{\widehat{PR}}$ and $\mathcal{A}$ families obtained the best results for the majority of the datasets. The results obtained with these two families are not significantly different. The $\mathcal{\widehat{C}}$ and $\mathcal{E}$ families have also similar results{, while both of them perform worse than  $\mathcal{\widehat{PR}}$ (and this difference is statistically significant).}
 
The lower performance of these families are due to the limited expressivity of their constraints. The performance of the $\mathcal{\widehat{NR}}$ family is close to the performance of $\mathcal{\widehat{C}}$ and $\mathcal{E}$ even {though} $\mathcal{\widehat{NR}}$ contains relation constraints. This is due to the lower discriminating power of the constraints in $\mathcal{\widehat{NR}}$ {that negate the occurrence of a target activity (when the activation occurs) rather than explicitly constraining the occurrence of a specific target activity (when the activation occurs) as for $\mathcal{\widehat{PR}}$ constraints. Positive relations (and existential constraints) seem hence  to contribute most to the good performance of our Prescriptive Process Monitoring system (\textbf{RQ2}).}

\revision{
The time efficiency and scalability performance are crucial for using the proposed system in a real environment that needs real-time performance. As the underlying DT can be learned offline, the time performance becomes relevant for the recommendation generation part (i.e., Algorithm~\ref{algo:RecommendationGeneration}). We therefore measured the time needed for generating the recommendations for prefixes $\sigma_k$ of different lengths. Figure~\ref{fig:rec_times} shows the time performance by comparing the case of the event log with the best performance (the $sepsis\_cases\_3$ in Figure~\ref{fig:sepsis_3_times}) with the one with the worst generation time (the $bpic2011\_4$ in Figure~\ref{fig:bpic11_4_times}).
\begin{figure}
     \centering
     \begin{subfigure}[b]{0.49\textwidth}
         \centering
         \includegraphics[width=\textwidth]{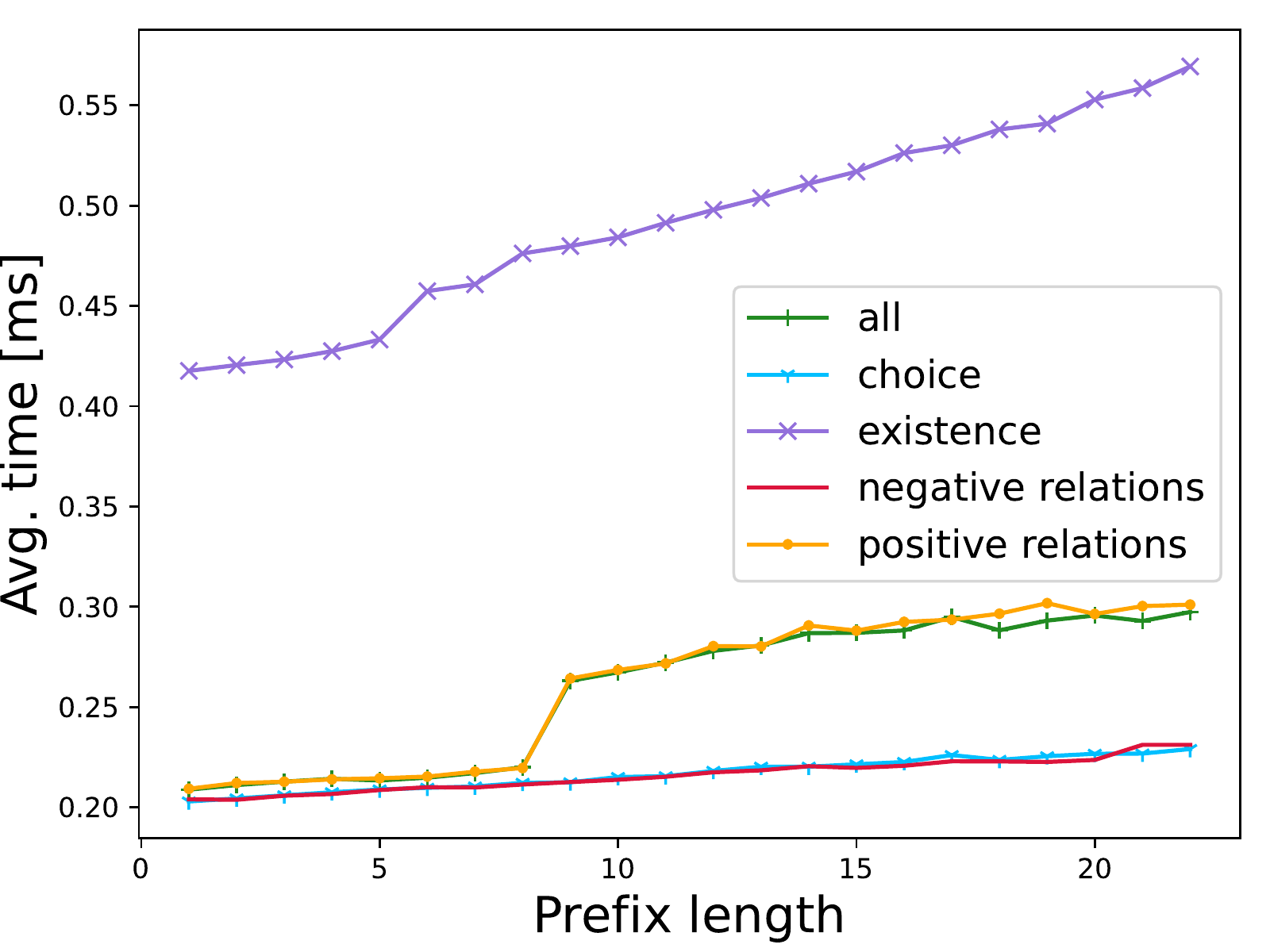}
         \caption{The $sepsis\_cases\_3$ event log has the best performance.}
         \label{fig:sepsis_3_times}
     \end{subfigure}
     \hfill
     \begin{subfigure}[b]{0.49\textwidth}
         \centering
         \includegraphics[width=\textwidth]{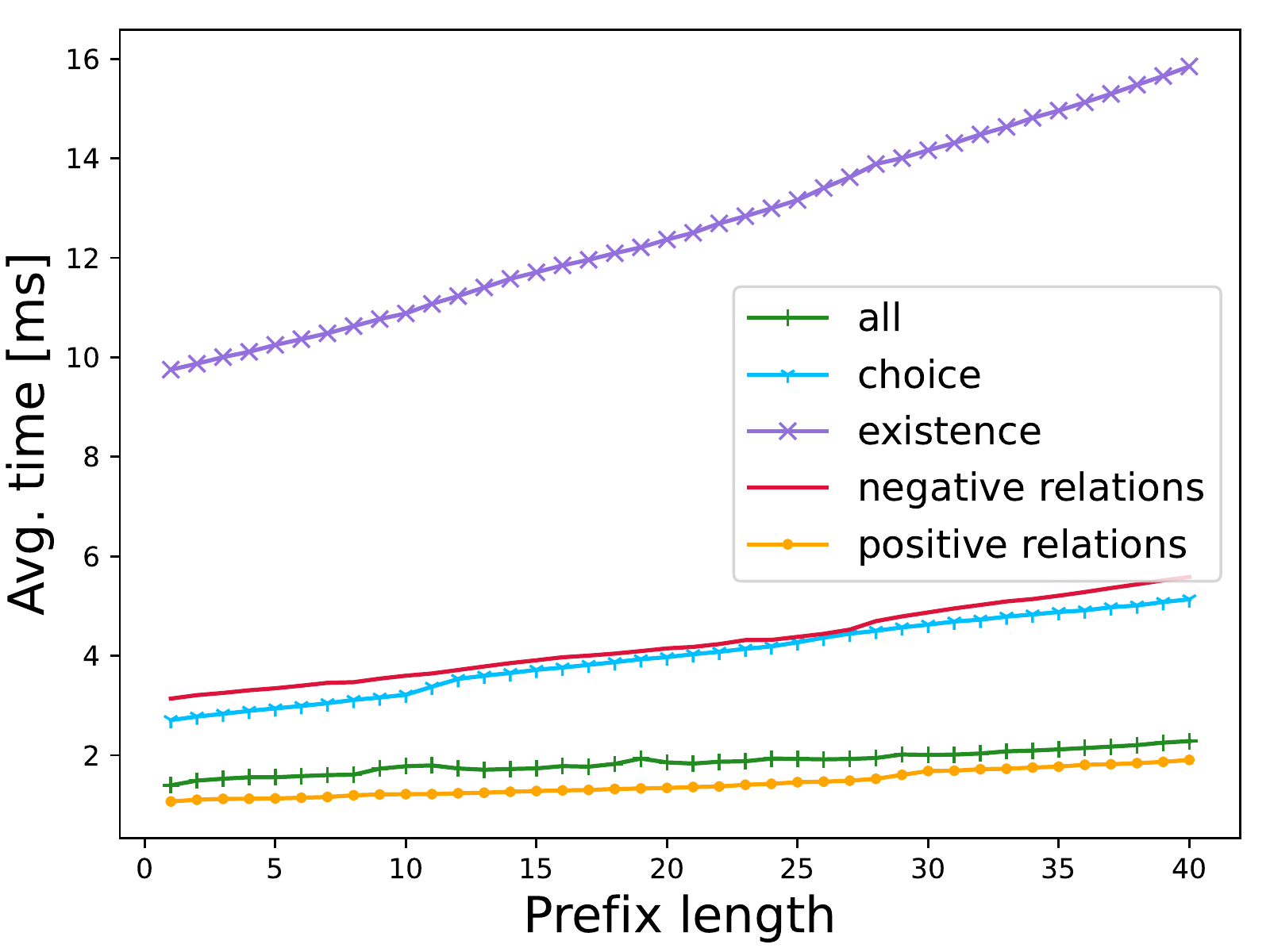}
         \caption{The $bpic2011\_4$ event log has the worst performance.}
         \label{fig:bpic11_4_times}
     \end{subfigure}
        \caption{The real-time scalability performance of the recommendations generation (few milliseconds) allow the system to be deployed in real settings.}
        \label{fig:rec_times}
\end{figure}
From the plot, we can see that our Prescriptive Process Monitoring system generates the recommendations in few milliseconds for both event logs. Such a high efficiency allows the system to be employed by process stakeholders to take decisions in real-time. The second insight we can draw from this experiment is that the generation time increases as the prefix length increases. This was expected since computing the RV satisfaction $[\sigma_k \models \varphi]_{RV}$ in the fitness function (Equations~\ref{eq:fitness} and \ref{eq:compliance}) linearly grows with the length of $\sigma_k$~\cite{DonadelloRMS22}. The time performance also depends on the number of paths of the DT and on their lengths (see lines 2 and 4 of Algorithm~\ref{algo:RecommendationGeneration}). In the case of both the considered event logs, the $\mathcal{E}$ \declare family in the log encoding produces larger DTs with respect to the other families and, therefore, requires more time for generating the recommendations.
}

\secRevision{
\subsection{Impact Analysis}
We discuss here the impact analysis~(similarly to the one performed in \cite{,math10071129}) of the proposed Prescriptive Process Monitoring system. The accuracy, tested on standard benchmarks of the process mining community, is satisfactory as shown in Table~\ref{tab:avg_fscore}: in most of the cases, following the prescriptions leads to a positive outcome of the ongoing case. In some cases ($production$ and $sepsis\_cases\_1$), our system does not have high performance. This is due to the fact that the \declare encoding focuses only on the control flow neglecting the information contained in the payload that could help in a better classification of the outcome and, therefore, in providing better recommendations. Different families of \declare constraints can be used for generating the recommendations. $\widehat{\mathcal{PR}}$ and $\mathcal{A}$ perform the best and are statistically better than $\mathcal{E}$ and $\widehat{\mathcal{C}}$ that use less expressive constraints. This is shown in Figure~\ref{fig:cd}. Our system has also good scalability performance (see Figure~\ref{fig:rec_times}) since, even with long prefixes, the computational times are in the order of milliseconds, thus making the system suitable for real-time applications. By construction, our system returns flexible (based on Linear Temporal Logic) and prioritized recommendations compared to the more static ones provided in other approaches. Table~\ref{tab:competitors} qualitatively compares our system with existing works. A quantitative comparison is not possible since other methods compute different types of recommendations that are not comparable with the ones of our proposal.
\begin{landscape}
\begin{table}[h!tbp]
\centering
\caption{Comparison between the proposed Prescriptive Process Monitoring system and the approaches prescribing recommendations on the control flow. CF indicates input features extracted from the control flow, T stands for features derived from the timestamps, FE indicates ad-hoc-engineered features, CA features computed from case attributes.}
\label{tab:competitors}
\begin{tabular}{@{}ccccc@{}}
\toprule
\textbf{Reference}                        & \textbf{\begin{tabular}[c]{@{}c@{}}Recommendation\\ target\end{tabular}} & \textbf{\begin{tabular}[c]{@{}c@{}}Input\\ features\end{tabular}} & \textbf{\begin{tabular}[c]{@{}c@{}}Modeling\\ technique\end{tabular}}             & \textbf{\begin{tabular}[c]{@{}c@{}}Output (type of\\ recommendations)\end{tabular}}                 \\ \midrule
Our   system                              & Categorical outcome                                                      & CF                                                               & \begin{tabular}[c]{@{}c@{}}Decision Trees and\\ Adaptive Querying\end{tabular}     & \begin{tabular}[c]{@{}c@{}}Linear Temporal Logic\\ Relationships among activities\end{tabular} \\
\cite{WeinzierlSZM20}    & Reducing defect rate                                                     & CF, FE                                                       & LSTM                                                                              & Next task to perform                                                                             \\
\cite{nakatumba2012meta} & Reducing cycle time                                                      & CF                                                               & Model-less                                                                         & Next task to perform                                                                             \\
\cite{LeoniDR20}         & Temporal outcome                                                         & FE                                                           & \begin{tabular}[c]{@{}c@{}}Random Forest,\\ SVM, Decision Trees\end{tabular}       & Next task to perform                                                                             \\
\cite{HeberHS15}         & Reducing cycle time                                                      & CF, T, FE                                                    & State Machine                                                                      & Next task to perform                                                                             \\
\cite{DetroSPLLB20}      & Reducing defect rate                                                     & FE, CA                                          & Decision Trees                                                                     & Set of tasks to perform                                                                          \\
\cite{BranchiFGM0R22}    & KPI maximization                                                         & CF, FE                                                       & Reinforcement Learning                                                             & Set of tasks to perform                                                                          \\
\cite{KotsiasKBTM22}     & Categorical outcome                                                      & CF                                                               & Reinforcement Learning                                                             & Next task to perform                                                                             \\
\cite{abs-2303-03572}    & Categorical outcome                                                      & FE                                                           & \begin{tabular}[c]{@{}c@{}}Reinforcement Learning\\ and Causal Forest\end{tabular} & Next task to perform                                                                             \\
\cite{WeinzierlDZM20}    & Temporal outcome                                                         & CF                                                               & LSTM                                                                              & Next task to perform                                                                             \\ \bottomrule
\end{tabular}
\end{table}
\end{landscape}
The first column of Table~\ref{tab:competitors} indicates the reference in which each approach was presented. The second column shows the goal that is supposed to be achieved with the prescribed recommendations. Our system provides recommendations for increasing the probability that an ongoing case ends with a positive (categorical) outcome. In addition, our system can easily deal with temporal outcomes (i.e., categorical outcomes derived from temporal information such as the violation of a planned cycle time or of a deadline) by simply changing the labeling function (see Definition~\ref{def:labeling}). Our approach does not deal with numerical labelings like the ones used in the approaches aiming at reducing the cycle time or the defect rate. Concerning the target of maximizing a certain KPI, this is tightly related to the concept of reward in Reinforcement Learning and the approach we propose is not (yet) able to work in such learning setting. Table~\ref{tab:competitors} also shows that almost all the other Prescriptive Process Monitoring systems use features coming from the control flow as our system does. As future work, we plan to use also features coming from data payloads. The most important difference between our system and other works is that our system is based on a totally novel type of recommendations. Other approaches, indeed, provide as recommendations a (set of) mandatory action(s) to perform next during an ongoing case, whereas our system provides a set of temporal relations to satisfy or violate in an ongoing case. This ensures flexibility without affecting the reliability of the system as shown in Table~\ref{tab:avg_fscore}. The type of recommendations provided by existing systems can be seen as a special case of the temporal relations provided by our system where the \declare encoding considers only the \constr{chain response} as \declare template and the recommendations prescribe to satisfy the logical relations
\constr{chain response} (\act{A\textsubscript{0}}, \act{A\textsubscript{1}}), \constr{chain response} (\act{A\textsubscript{1}}, \act{A\textsubscript{2}}) ... \constr{chain response} (\act{A\textsubscript{n}}, \act{A\textsubscript{n+1}}), where \act{A\textsubscript{0}} is the last activity of the prefix $\sigma_k$ of the ongoing case.}

\subsection{Limitations}
\label{sec:limitations}

The main limitation of our work relates to the fact that our Prescriptive Process Monitoring system has been evaluated in an \emph{offline scenario}. In particular, we evaluated our system by performing a ``what-if'' analysis. We tried to mitigate this limitation by running our experiments using different real-life datasets. However, for a final deployment in a real organization, our system {would} require further evaluations with real users {employing} the system in their worklife. This requires a user-friendly Graphical User Interface (GUI) that allows users to interact with the system. 
To this aim, in the future, we plan to embed our recommendation system in the Nirdizati tool~\cite{RizziSFGKM19}, which is an open-source web-based Predictive Process Monitoring engine. Moreover, in real scenarios, it could happen that users are not familiar with \declare constraints. Therefore, a human-understandable rendering of the prescriptions could be more effective. This could be achieved with the use of Natural Language Generation techniques for persuasive messages~\cite{DragoniDE20}, where recommendations are passed as input. The resulting persuasive natural language sentences contain an effective description of the constraints to satisfy, their importance for the achievement of a positive outcome of the process and some explanations to motivate them.

\revision{Another limitation is that the considered encoding is based on standard \declare patterns, i.e., on pure control flow features. These features do not take into account the data payloads as, for instance, resources. The data payloads can be injected in the encoding in a principled way since the \declare language has already been extended to include data conditions \cite{BurattinMS16}. This would improve the system performance in those cases in which the only control flow constraints are not sufficient for an effective outcome-based discrimination of positive and negative cases, such as for the $hospital\_billing\_2$ and the $sepsis\_case\_1$ datasets (see Section~\ref{sec:results}).}


\section{Conclusion}
\label{sec:conclusion}

\par{\textbf{Findings.} The proposed Outcome-Oriented Prescriptive Process Monitoring approach aims at providing recommendations to maximize the likelihood a positive process outcome. Differently from state-of-the-art works, the proposed approach does not recommend specific activities to be executed, but temporal properties among activities that need to be preserved or violated. This type of recommendations avoid forcing the execution of specific activities during the process execution thus providing more flexibility in the ways the process should be executed. The approach has been evaluated on a pool of 22 real-life event logs already used as a benchmark in Predictive Process Monitoring in  \cite{TeinemaaDRM19}. \revision{For most of the datasets (18 out of 22), we achieved an F1 score higher than 90\%. Our proposal is efficient as the generation of the recommendations for an ongoing case can be performed in few milliseconds. Therefore, our system can support process stakeholders that need to intervene on an ongoing case to have a positive outcome in real-time.}}


\par{\textbf{Limitations.} \revision{The evaluation we proposed is mainly based on a ``what-if'' analysis, that is, on simulations with real-life event logs. However, to have a system that can be deployed in a real organization, an evaluation with real process stakeholders is necessary. This would test both the reliability of the recommendations in a real working scenario and their intelligibility to real users. The other main limitation of our system is that the features used for encoding an event log are pure control flow \declare patterns in which data payloads are neglected. This limits the expressive power of the Machine Learning model used to generate the recommendations. Another limitation is that, in the current implementation, our system provides recommendations to achieve only categorical outcomes neglecting numeric outcomes like, e.g., the cycle time of a case.}}

\par{\textbf{Future work.} \revision{The future work will address the above limitations. We plan to extend our current evaluation by deploying the proposed approach in an organization environment and by performing experiments with real users. This would allow us to understand whether the recommendations provided by our system are useful and comprehensible to process analysts as similarly done in~\cite{GalantiLMNMSM23} in the context of model explanations. In addition, we plan i) to use a richer event-log encoding that uses data payloads for building the features, and ii) to extend the current implementation to other type of outcomes/prediction tasks like, for example, the regression-based ones.
We are also interested in investigating more advanced Machine Learning models for learning the correlations between \declare constraints and the outcome of a trace, such as Neural Networks paired with Explainable AI techniques~\cite{0001WBM20}. Lastly, future directions will involve the use of causal effect estimations~\cite{BOZORGI2023102198} to understand the impact on the ongoing case of respecting or not the prescribed recommendations.}}



\bibliographystyle{elsarticle-num} 
\bibliography{bibliography}

\begin{thebibliography}{10}
\expandafter\ifx\csname url\endcsname\relax
  \def\url#1{\texttt{#1}}\fi
\expandafter\ifx\csname urlprefix\endcsname\relax\def\urlprefix{URL }\fi
\expandafter\ifx\csname href\endcsname\relax
  \def\href#1#2{#2} \def\path#1{#1}\fi

\bibitem{TeinemaaTLDM18}
I.~Teinemaa, N.~Tax, M.~{de Leoni}, M.~Dumas, F.~M. Maggi, Alarm-based
  prescriptive process monitoring, in: {BPM} (Forum), Vol. 329 of Lecture Notes
  in Business Information Processing, Springer, 2018, pp. 91--107.

\bibitem{DBLP:journals/kais/Fahrenkrog-Petersen22}
S.~A. Fahrenkrog{-}Petersen, N.~Tax, I.~Teinemaa, M.~Dumas, M.~{de Leoni},
  F.~M. Maggi, M.~Weidlich, Fire now, fire later: alarm-based systems for
  prescriptive process monitoring, Knowl. Inf. Syst. 64~(2) (2022) 559--587.

\bibitem{Aalst16book}
W.~M.~P. van~der Aalst, Process Mining - Data Science in Action, Second
  Edition, Springer, 2016.

\bibitem{PeSV07}
M.~Pesic, H.~Schonenberg, W.~M.~P. van~der Aalst, {DECLARE:} full support for
  loosely-structured processes, in: {EDOC}, {IEEE} Computer Society, 2007, pp.
  287--300.

\bibitem{WeinzierlSZM20}
S.~Weinzierl, M.~Stierle, S.~Zilker, M.~Matzner, A next click recommender
  system for web-based service analytics with context-aware lstms, in: {HICSS},
  ScholarSpace, 2020, pp. 1--10.

\bibitem{LeoniDR20}
M.~{de Leoni}, M.~Dees, L.~Reulink, Design and evaluation of a process-aware
  recommender system based on prescriptive analytics, in: {ICPM}, {IEEE}, 2020,
  pp. 9--16.

\bibitem{BranchiFGM0R22}
S.~Branchi, C.~{di Francescomarino}, C.~Ghidini, D.~Massimo, F.~Ricci,
  M.~Ronzani, Learning to act: {A} reinforcement learning approach to recommend
  the best next activities, in: {BPM} (Forum), Vol. 458 of Lecture Notes in
  Business Information Processing, Springer, 2022, pp. 137--154.

\bibitem{DBLP:journals/jodsn/CiccioM015}
C.~{Di Ciccio}, A.~Marrella, A.~Russo, Knowledge-intensive processes:
  Characteristics, requirements and analysis of contemporary approaches, J.
  Data Semant. 4~(1) (2015) 29--57.

\bibitem{TeinemaaDRM19}
I.~Teinemaa, M.~Dumas, M.~{la Rosa}, F.~M. Maggi, Outcome-oriented predictive
  process monitoring: Review and benchmark, {ACM} Trans. Knowl. Discov. Data
  13~(2) (2019) 17:1--17:57.

\bibitem{Kubrak21Quovadis}
K.~Kubrak, F.~Milani, A.~Nolte, M.~Dumas, Prescriptive process monitoring:
  \emph{Quo vadis}?, PeerJ Comput. Sci. 8 (2022) e1097.

\bibitem{nakatumba2012meta}
J.~Nakatumba, M.~Westergaard, W.~M. van~der Aalst, A meta-model for operational
  support, BPM Center Report BPM-12-05, BPMcenter. org (2012) 16--32.

\bibitem{HeberHS15}
E.~Heber, H.~Hagen, M.~Schmollinger, Application of process mining for
  improving adaptivity in case management systems, in: {DEC}, Vol. {P-244} of
  {LNI}, {GI}, 2015, pp. 221--231.

\bibitem{DetroSPLLB20}
S.~P. Detro, E.~A.~P. Santos, H.~Panetto, E.~D. F.~R. Loures, M.~Lezoche, C.~M.
  C.~M. Barra, Applying process mining and semantic reasoning for process model
  customisation in healthcare, Enterp. Inf. Syst. 14~(7) (2020) 983--1009.

\bibitem{KotsiasKBTM22}
S.~Kotsias, A.~Kerasiotis, A.~Bousdekis, G.~Theodoropoulou, G.~Miaoulis,
  Predictive and prescriptive business process monitoring with reinforcement
  learning, in: NiDS, Vol. 556 of Lecture Notes in Networks and Systems,
  Springer, 2022, pp. 245--254.

\bibitem{abs-2303-03572}
Z.~D. Bozorgi, M.~Dumas, M.~{la Rosa}, A.~Polyvyanyy, M.~Shoush, I.~Teinemaa,
  Learning when to treat business processes: Prescriptive process monitoring
  with causal inference and reinforcement learning, CoRR abs/2303.03572 (2023).

\bibitem{WibisonoNBP15}
A.~Wibisono, A.~S. Nisafani, H.~Bae, Y.~Park, On-the-fly performance-aware
  human resource allocation in the business process management systems
  environment using na{\"{\i}}ve bayes, in: {AP-BPM}, Vol. 219 of Lecture Notes
  in Business Information Processing, Springer, 2015, pp. 70--80.

\bibitem{SindhgattaGD16}
R.~Sindhgatta, A.~K. Ghose, H.~K. Dam, Context-aware analysis of past process
  executions to aid resource allocation decisions, in: CAiSE, Vol. 9694 of
  Lecture Notes in Computer Science, Springer, 2016, pp. 575--589.

\bibitem{Yaghoibi2017Cycle}
M.~Yaghoibi, M.~Zahedi, Cycle time reduction and runtime rebalancing by
  reallocating dependent tasks, International Journal of Engineering 30~(12)
  (2017) 1831--1839.

\bibitem{abdulhameed2018resource}
N.~Abdulhameed, I.~Helal, A.~Awad, E.~Ezat, A resource recommendation approach
  based on co-working history, Int. J. Adv. Comput. Sci. Appl. 9~(7) (2018)
  236--245.

\bibitem{ShoushD21}
M.~Shoush, M.~Dumas, Prescriptive process monitoring under resource
  constraints: {A} causal inference approach, in: {ICPM} Workshops, Vol. 433 of
  Lecture Notes in Business Information Processing, Springer, 2021, pp.
  180--193.

\bibitem{NezhadB11}
H.~R.~M. Nezhad, C.~Bartolini, Next best step and expert recommendation for
  collaborative processes in {IT} service management, in: {BPM}, Vol. 6896 of
  Lecture Notes in Computer Science, Springer, 2011, pp. 50--61.

\bibitem{BarbaWV11}
I.~Barba, B.~Weber, C.~D. Valle, Supporting the optimized execution of business
  processes through recommendations, in: Business Process Management Workshops
  {(1)}, Vol.~99 of Lecture Notes in Business Information Processing, Springer,
  2011, pp. 135--140.

\bibitem{BozorgiTDRP21}
Z.~D. Bozorgi, I.~Teinemaa, M.~Dumas, M.~{la Rosa}, A.~Polyvyanyy, Prescriptive
  process monitoring for cost-aware cycle time reduction, in: {ICPM}, {IEEE},
  2021, pp. 96--103.

\bibitem{YangDSZFXBM17}
S.~Yang, X.~Dong, L.~Sun, Y.~Zhou, R.~A. Farneth, H.~Xiong, R.~S. Burd,
  I.~Marsic, A data-driven process recommender framework, in: {KDD}, {ACM},
  2017, pp. 2111--2120.

\bibitem{BOZORGI2023102198}
Z.~D. Bozorgi, I.~Teinemaa, M.~Dumas, M.~{la Rosa}, A.~Polyvyanyy, Prescriptive
  process monitoring based on causal effect estimation, Information Systems
  (2023) 102198.

\bibitem{Thomas2017Recommending}
L.~Thomas, M.~M. Kumar, B.~Annappa, Recommending an alternative path of
  execution using an online decision support system, in: Proceedings of the
  2017 international conference on intelligent systems, metaheuristics \& swarm
  intelligence, 2017, pp. 108--112.

\bibitem{metzger2019proactive}
A.~Metzger, A.~Neubauer, P.~Bohn, K.~Pohl, Proactive process adaptation using
  deep learning ensembles, in: Int. Conf. on Advanced Information Systems
  Engineering, Springer, 2019, pp. 547--562.

\bibitem{MetzgerKP20}
A.~Metzger, T.~Kley, A.~Palm, Triggering proactive business process adaptations
  via online reinforcement learning, in: {BPM}, Vol. 12168 of Lecture Notes in
  Computer Science, Springer, 2020, pp. 273--290.

\bibitem{ShoushD22}
M.~Shoush, M.~Dumas, When to intervene? prescriptive process monitoring under
  uncertainty and resource constraints, in: {BPM} (Forum), Vol. 458 of Lecture
  Notes in Business Information Processing, Springer, 2022, pp. 207--223.

\bibitem{DeVa13}
G.~{de Giacomo}, M.~Y. Vardi, Linear temporal logic and linear dynamic logic on
  finite traces, in: {IJCAI}, {IJCAI/AAAI}, 2013, pp. 854--860.

\bibitem{PEEPERKORN2023106393}
J.~Peeperkorn, S.~vanden Broucke, J.~De~Weerdt, Global conformance checking
  measures using shallow representation and deep learning, Engineering
  Applications of Artificial Intelligence 123 (2023) 106393.

\bibitem{Beer01efficientdetection}
I.~Beer, S.~Ben{-}David, C.~Eisner, Y.~Rodeh, Efficient detection of vacuity in
  temporal model checking, Formal Methods Syst. Des. 18~(2) (2001) 141--163.

\bibitem{kupf:vacu03}
O.~Kupferman, M.~Y. Vardi, Vacuity detection in temporal model checking, Int.
  J. Softw. Tools Technol. Transf. 4~(2) (2003) 224--233.

\bibitem{Pesic2008}
M.~Pesic, Constraint-based workflow management systems: Shifting control to
  users, Ph.D. thesis, Industrial Engineering and Innovation Sciences,
  proefschrift. (2008).
\newblock \href {https://doi.org/10.6100/IR638413}
  {\path{doi:10.6100/IR638413}}.

\bibitem{DBLP:conf/edoc/MaggiMB19}
F.~M. Maggi, M.~Montali, U.~Bhat, Compliance monitoring of multi-perspective
  declarative process models, in: {EDOC}, {IEEE}, 2019, pp. 151--160.

\bibitem{agrawalApriori}
R.~Agrawal, R.~Srikant, Fast algorithms for mining association rules in large
  databases, in: {VLDB}, Morgan Kaufmann, 1994, pp. 487--499.

\bibitem{ross2014mutual}
B.~C. Ross, Mutual information between discrete and continuous data sets, PloS
  one 9~(2) (2014) e87357.

\bibitem{pnueli1977temporal}
A.~Pnueli, The temporal logic of programs, in: {FOCS}, {IEEE} Computer Society,
  1977, pp. 46--57.

\bibitem{Dumas21}
M.~Dumas, Constructing digital twins for accurate and reliable what-if business
  process analysis, in: Problems@BPM, Vol. 2938 of {CEUR} Workshop Proceedings,
  CEUR-WS.org, 2021, pp. 23--27.

\bibitem{DonadelloRMS22}
I.~Donadello, F.~Riva, F.~M. Maggi, A.~Shikhizada, Declare4py: {A} python
  library for declarative process mining, in: {BPM} (PhD/Demos), Vol. 3216 of
  {CEUR} Workshop Proceedings, CEUR-WS.org, 2022, pp. 117--121.

\bibitem{scikit-learn}
F.~Pedregosa, G.~Varoquaux, A.~Gramfort, V.~Michel, B.~Thirion, O.~Grisel,
  M.~Blondel, P.~Prettenhofer, R.~Weiss, V.~Dubourg, J.~Vanderplas, A.~Passos,
  D.~Cournapeau, M.~Brucher, M.~Perrot, E.~Duchesnay, Scikit-learn: Machine
  learning in {P}ython, Journal of Machine Learning Research 12 (2011)
  2825--2830.

\bibitem{LeontjevaCFDM15}
A.~Leontjeva, R.~Conforti, C.~{di Francescomarino}, M.~Dumas, F.~M. Maggi,
  Complex symbolic sequence encodings for predictive monitoring of business
  processes, in: {BPM}, Vol. 9253 of Lecture Notes in Computer Science,
  Springer, 2015, pp. 297--313.

\bibitem{dfmMaking2022}
C.~{di Francescomarino}, I.~Donadello, C.~Ghidini, F.~M. Maggi, W.~Rizzi,
  Making sense of temporal data: the {DECLARE} encoding, in: PMAI@IJCAI, Vol.
  3310 of {CEUR} Workshop Proceedings, CEUR-WS.org, 2022, pp. 77--80.

\bibitem{JamaludinRKBMM22}
S.~Z.~M. Jamaludin, N.~A. Romli, M.~S.~M. Kasihmuddin, A.~Baharum, M.~A.
  Mansor, M.~F. Marsani, Novel logic mining incorporating log linear approach,
  J. King Saud Univ. Comput. Inf. Sci. 34~(10 Part {B}) (2022) 9011--9027.

\bibitem{math10060915}
M.~S.~M. Kasihmuddin, S.~Z.~M. Jamaludin, M.~A. Mansor, H.~A. Wahab, S.~M.~S.
  Ghadzi, Supervised learning perspective in logic mining, Mathematics 10~(6)
  (2022).

\bibitem{Demsar06}
J.~Demsar, Statistical comparisons of classifiers over multiple data sets, J.
  Mach. Learn. Res. 7 (2006) 1--30.

\bibitem{math10071129}
S.~S. Muhammad~Sidik, N.~E. Zamri, M.~S. Mohd~Kasihmuddin, H.~A. Wahab, Y.~Guo,
  M.~A. Mansor, Non-systematic weighted satisfiability in discrete hopfield
  neural network using binary artificial bee colony optimization, Mathematics
  10~(7) (2022).

\bibitem{WeinzierlDZM20}
S.~Weinzierl, S.~Dunzer, S.~Zilker, M.~Matzner, Prescriptive business process
  monitoring for recommending next best actions, in: {BPM} (Forum), Vol. 392 of
  Lecture Notes in Business Information Processing, Springer, 2020, pp.
  193--209.

\bibitem{RizziSFGKM19}
W.~Rizzi, L.~Simonetto, C.~{di Francescomarino}, C.~Ghidini, T.~Kasekamp, F.~M.
  Maggi, Nirdizati 2.0: New features and redesigned backend, in: {BPM}
  (PhD/Demos), Vol. 2420 of {CEUR} Workshop Proceedings, CEUR-WS.org, 2019, pp.
  154--158.

\bibitem{DragoniDE20}
M.~Dragoni, I.~Donadello, C.~Eccher, Explainable {AI} meets persuasiveness:
  Translating reasoning results into behavioral change advice, Artif. Intell.
  Medicine 105 (2020) 101840.

\bibitem{BurattinMS16}
A.~Burattin, F.~M. Maggi, A.~Sperduti, Conformance checking based on
  multi-perspective declarative process models, Expert Syst. Appl. 65 (2016)
  194--211.

\bibitem{GalantiLMNMSM23}
R.~Galanti, M.~{de Leoni}, M.~Monaro, N.~Navarin, A.~Marazzi, B.~D. Stasi,
  S.~Maldera, An explainable decision support system for predictive process
  analytics, Eng. Appl. Artif. Intell. 120 (2023) 105904.

\bibitem{0001WBM20}
R.~Confalonieri, T.~Weyde, T.~R. Besold, F.~M. del Prado~Mart{\'{\i}}n,
  {TREPAN} reloaded: {A} knowledge-driven approach to explaining black-box
  models, in: {ECAI}, Vol. 325 of Frontiers in Artificial Intelligence and
  Applications, {IOS} Press, 2020, pp. 2457--2464.

\end{thebibliography}

\end{document}